\pdfoutput=1

\newif\ifreview
\reviewfalse

\ifreview
\documentclass[review]{elsarticle} 
\else
\documentclass[5p,twocolumn]{elsarticle}
\fi

\usepackage{hyperref}
\usepackage{amsmath} 
\usepackage{amssymb}  
\usepackage{pifont}
\usepackage{siunitx}
\usepackage{caption}
\usepackage{subcaption}
\usepackage{graphicx}
\usepackage{makecell}
\usepackage{booktabs}
\usepackage{units}
\usepackage{balance}
\usepackage{multirow}
\usepackage[table]{xcolor}
\usepackage{array}
\usepackage{tikz}
\usepackage{capt-of}
\usepackage[nolist,nohyperlinks]{acronym}

\journal{Automation in Construction}

\begin{document}


\begin{frontmatter}

\title{HEAP - The Autonomous Walking Excavator\tnoteref{t1}}
\tnotetext[t1]{This work was supported in part by the Swiss National Science Foundation through the National Centre of Competence in Digital Fabrication (NCCR dfab), Hexagon Geosystems and armasuisse Science and Technology.}

\author[eth]{Dominic Jud\corref{cor1}}
\ead{dominic.jud@mavt.ethz.ch}
\author[eth]{Simon Kerscher}
\author[eth]{Martin Wermelinger}
\author[eth]{Edo Jelavic}   
\author[eth]{Pascal Egli}
\author[moog]{Philipp Leemann}
\author[eth]{Gabriel Hottiger}
\author[eth]{Marco Hutter}
\ead{mahutter@ethz.ch}
\address[eth]{Robotic Systems Lab, ETH Zurich, Switzerland}
\address[moog]{Moog Industrial Group GmbH, B\"oblingen, Germany}
\cortext[cor1]{Corresponding author}

\begin{abstract}
The demand and the potential for automation in the construction sector is unmatched, particularly for increasing environmental sustainability, improving worker safety and reducing labor shortages. We have developed an autonomous walking excavator - based one of the most versatile machines found on construction sites - as one way to begin fulfilling this potential. This article describes the process of converting an off-the-shelf construction machine into an autonomous robotic system. First we outline the necessary sensing equipment for full autonomy and the novel actuation of the legs, and compare three different complementary actuation principles for the excavator's arm. Second, we solve the state estimation problem for a general wheeled-legged robot. Beside kinematic measurements, it includes GNSS-RTK, to absolutely reference the machine on a construction site. Third, we developed individual controllers for driving, chassis balancing and arm motions allowing for fully autonomous operation. Lastly, we highlight the machine's potential in four different real-world applications, e.g. autonomous trench digging, autonomous assembly of dry stone walls, autonomous forestry work and semi-autonomous teleoperation. On top, we also share some development insights and possible future research directions.
\end{abstract}


\begin{acronym}
\acro{DoF}{degrees of freedom}
\acro{LiDAR}{light detection and ranging}
\acro{GPS}{Global Positioning System}
\acro{ICP}{iterative closest point}
\acro{MPC}{model predictive control}
\acro{RTK}{real-time kinematic}
\acro{IMU}{inertial measurement unit}
\acro{GNSS}{global navigation satellite system}
\acro{ROS}{Robot Operating System}
\acro{CAD}{computer aided design}
\acro{IF}{In-Situ Fabricator}
\acro{ICM}{Integrated Control Module}
\end{acronym}

\end{frontmatter}


\section{Introduction}

\subsection{Motivation}

The degree of automation in the construction sector remains low, despite great potential and demand \cite{Chui2019}. In contrast, other sectors like finance, information and communication, have undergone a massive transformations in the last decades, predominantly pushed by advances in data analysis and processing. Achieving a similar level of transformation in construction seems more challenging, as it requires automation of physical machines into dexterous and autonomous robots that can cope with real-world problems \cite{Hawksworth2018}. Nevertheless, such automation and digitization has the potential to solve some of the most pressing problems in construction, agriculture, forestry and related sectors. By increasing efficiency and fostering more sustainable techniques, automation could help these sectors become more environmentally friendly. On-site safety will increase when the workers are relieved from the most dangerous and dirty jobs \cite{Petersen19}, \cite{Ardiny2015}. And potential labor shortages can also be reduced as demographic change, fewer young people willing to work in these industries and competition from a growing service sector makes this a pressing matter especially in the Asian part of the world \cite{Dobbs2014}. 

\begin{figure}
\centering
\includegraphics[width=\linewidth]{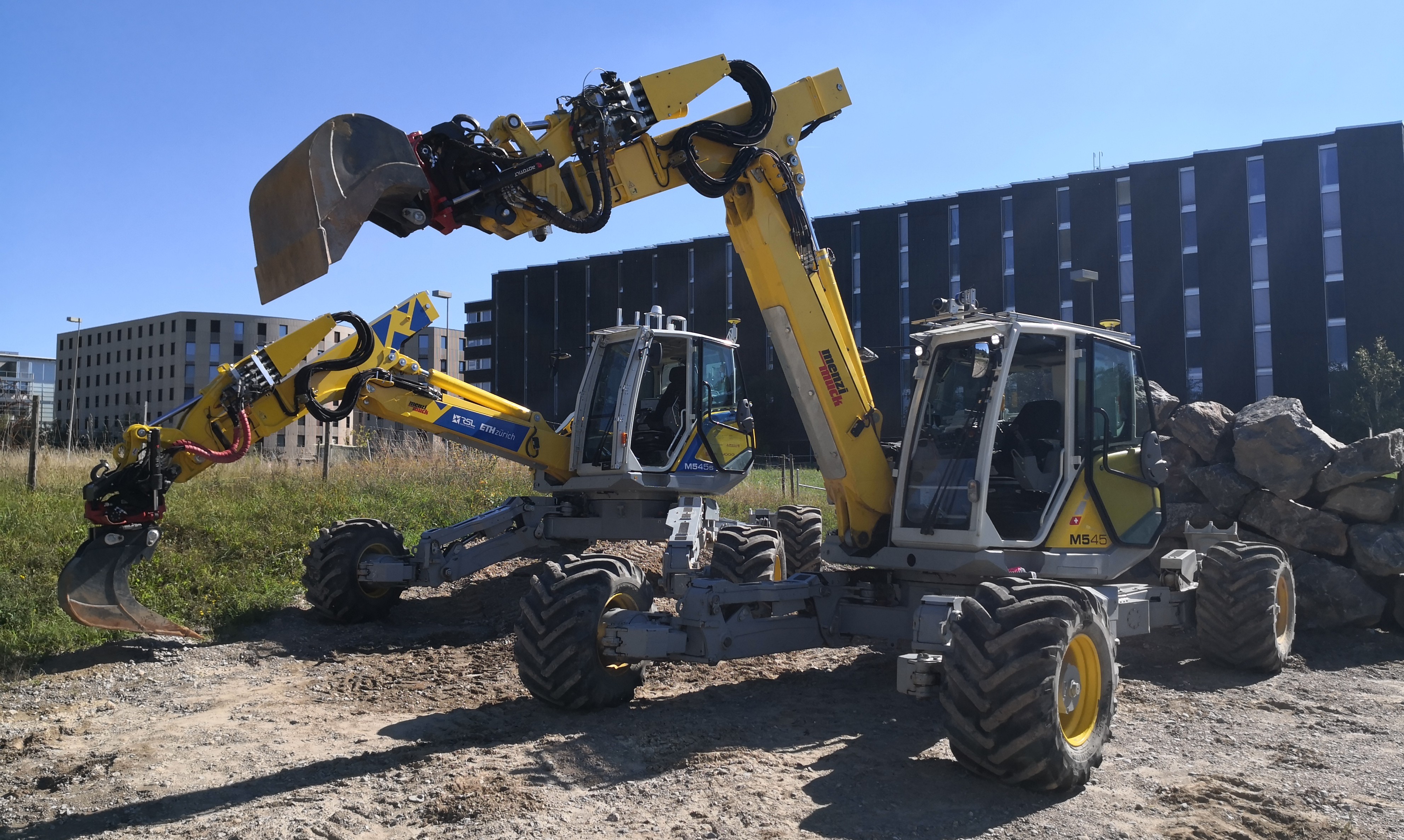}
\caption{HEAP, the first of its kind for autonomous deployment, in front of its clone Armano, a proof of concept for teleoperated/semi-autonomous operation.}
\label{fig:heapAndArmasuisse}
\vspace{-0.5cm}
\end{figure}

An on-site robot can be a new concept from the ground up, or an already existing human-operated machine can be transformed into an autonomous agent \cite{Melenbrink2020}. In the latter case, already existing knowledge and technologies lead to shorter development times and lower costs. The robots built from existing machines can then be categorized into stationary and mobile systems. While stationary robots, e.g. gantry, crane or cable robots \cite{Bruckmann18}, are relatively easy to automate, they can only build structures that fit in their workspace. Mobile robots, on the other hand, can build structures larger than their own size. However, the mobility aspect introduces multiple additional challenges such as state estimation, localization, environment perception, motion planning and control. These challenges can be particularly hard for mobile robots on construction sites, as these environments are notoriously unstructured and quickly changing. 

As a proof of feasibility and to push the state of the art in construction machine automation, we set out to automate the most versatile existing mobile machine on construction sites, a walking excavator (chosen over tracked or wheeled excavators due to its superior mobility). We worked with multiple industry partners for over half a decade to bring this project to fruition.  This article introduces the autonomous walking excavator HEAP (\textbf{H}ydraulic \textbf{E}xcavator for an \textbf{A}utonomous \textbf{P}urpose) shown in Figure \ref{fig:heapAndArmasuisse} and highlights its potential for automation in various applications within the construction industry.

\subsection{Related Work}

\subsubsection{Related Construction Robots}
So far, mobile robots for construction in research are usually built from a skid-steer wheeled or tracked mobile base with an industrial arm placed on top \cite{Doerfler19}, \cite{Gawel2019}. These industrial arms were initially built for static use in manufacturing and assembly lines. In contrast to hydraulic construction machines, they are optimized for precision and typically suffer from a low payload to weight ratio, resulting in a relatively heavy and weak mobile manipulator. Additionally, the skid-steer base only allows them to move on relatively flat terrain. Giftthaler et al. \cite{giftthaler2017} propose an improvement of their traditional robot \ac{IF}, a tracked skid-steer base with an industrial arm, that features four wheeled legs and one or more arms, called \ac{IF}2. The use of hydraulic actuation would yield a better payload to weight ratio and the wheeled-legged base improves maneuverability and stability. When comparing the kinematic particularities of the visionary \ac{IF}2 to a much larger walking excavator, they are remarkably similar: A walking excavator is also equipped with four wheeled legs with three degrees of freedom per leg and an arm for manipulation. 

In their ROBDEKON project, Petereit et al. \cite{Petereit2019} suggest using a walking excavator for decommissioning tasks and excavation of contaminated soil. This proposal is not yet realized, but according to the available information, they identified walking excavators as an ideal solution for their application due to the machine's capabilities to traverse rough terrain. The initial experience was gathered in building up IOSB.BoB \cite{Emter2017}, a small tracked excavator that can dig simple trenches autonomously and can also be teleoperated. There is no other scientific work on automating walking excavators. However, there has been practical research over decades on automating standard tracked and wheeled excavators for excavation work, as well as relevant research on the automation of smaller hydraulic arms, e.g. originally remote-controlled demolition robots (Brock or similar).

\subsubsection{Automation of Hydraulic Excavators}
Historically, operators moved the cylinders of an excavator through levers directly attached to the valves. For improved sensitivity and less operator fatigue, a low pressure hydraulic pilot stage was introduced to drive the large main valves. This is still the most common way excavators are built today. The challenge in automation is that the valve characteristics are chosen to fit human operators and thus, for safety reasons, they have a large valve overlap and dead band. With this in mind, there are three meaningful ways of automate an excavator. 

First, the most radical way would be to remove all valves and replace them with high-performance servo valves. The project at the Australian Centre for Field Robotics (ACFR) by Quang et al. \cite{Quang2002} is a small 1.5\si{\tonne} excavator with electrohydraulic servo valves in the main stage and completely removed operator controls and pilot stage. Servo valves in larger machines have not been realized so far, mainly due to the high oil flow required for larger machines, which would imply large, costly servo valves. 

Second, this problem is often mitigated for larger machines by installing electric pilot stage valves to utilize the existing main valves, e.g., as used in the 25\si{\tonne} Autonomous Loading System (ALS) by Cannon et al. \cite{Cannon1999}. They use electric pilot stage valves to create an excavator to load trucks autonomously. The same setup was also put in place by Haga et al. \cite{Haga2001}, Groll et al. \cite{Groll2019} and Yang et al. \cite{Yang2019}. Tafazoli et al. \cite{Tafazoli2002} also modified the pilot stage of a mini excavator to control the arm by adding a pair of ON/OFF solenoid valves operated with differential pulse width modulation. This strategy turns an excavator into a steer-by-wire system. Similar to the trend in cars, excavator manufacturers are now starting to install electric pilot stage valves in their machines. This allows for operator specific characteristics of the joysticks and many more features, but will also enable automation of these machines without changing the hydraulics.

Third, different projects have also investigated the actuation of operator joysticks to automate an excavator's arm. Sasaki et al. \cite{Sasaki2008} used two pneumatic robot arms for this task. A more straightforward approach with only one degree of freedom per actuated joystick axis was chosen by Shin et al. \cite{Shin2012}. However, the mechanism is installed above the joysticks, which prevents an operator from entering the cabin, similar to the pneumatic arm. Lee et al. \cite{Lee2015} show a setup where the two motors for one joystick are mounted without preventing the operator from using the machine with the system in place. 

\subsubsection{Wheeled-Legged Robots}

A mobile robot needs to know its 6-dof pose and velocity for autonomous missions. This state estimation problem is traditionally solved by using an \ac{IMU} in conjunction with some localization sensor. For the In-Situ Fabricator, a \ac{LiDAR} and an \ac{IMU} are fused in a non-linear least-squares optimization \cite{Sandy2016}. All wheeled-legged robots still use this minimal setup. Examples are the mars rover MAMMOTH \cite{Reid2016} which relies on an iterative closest point (ICP) method and Momaro \cite{Schwarz2016} which fuses a rotating line laser with an \ac{IMU}. Although wheeled-legged robots could use their legs and wheels as odometry, no such work has been done so far. For legged robots with point feet, this has been known as legged odometry and is used on many platforms \cite{imulegkin}. \cite{Bjelonic2019} uses the implementation shown in this article, but only fuses data from an \ac{IMU} with kinematic measurements for dead reckoning without \ac{GNSS} or \ac{LiDAR} localization.

\subsection{Contribution and organization of this article}
In this article, we present HEAP, an autonomous multi-purpose mobile manipulator based on a walking excavator, which we deployed and tested in multiple real-world applications.

In Section \ref{sec:system_description}, the system is described with the installed sensors and the different actuation concepts that were realized and evaluated for the arm and the unique chassis. The three most commonly used ways of automating an excavator's arm are realized and compared, i.e. electric pilot stage, actuated joystick and servo valves. The novel joystick actuation setup is minimal (one motor per joystick axis), but still allows the operator to enter the cabin and use the joysticks and additionally improves the efficiency of the actuation through shorter lever arms compared to Lee et al. \cite{Lee2015}.

In Section \ref{sec:state_estimation}, a state estimation approach for general wheeled-legged robots is derived, comprised of \ac{IMU} predictions, \ac{GNSS} \ac{RTK} updates (or any other positioning source), rolling predictions from the wheels and legged odometry updates. This is the first state estimation formulation for wheeled-legged robots using leg kinematics and wheel odometry. It is evaluated especially for the use case on a walking excavator.

The chosen control approaches for driving, chassis balancing and arm motions are shown in Section \ref{sec:control}. The unique self-balancing chassis has shown its reliability in hundreds of hours of operation during autonomous experiments, as well as teleoperated missions. Slopes as steep as 100\%, deep mud and rough terrains were easily handled without any failures. Together with the inverse kinematic and inverse dynamic arm controllers that include important limits, they lay the foundation for unmanned operation.

In Section \ref{sec:applications}, we highlight four exemplary applications of HEAP in autonomous construction tasks that build upon the contributions from Sections \ref{sec:system_description}-\ref{sec:control}. 

Finally, some integration insights gathered during the conversion of a walking excavator to an autonomous robot are summarized in Section \ref{sec:lessons_learned} and a conclusion on this work with an outlook to future research directions is given in Section \ref{sec:conclusion}.

\section{System Description} \label{sec:system_description}

\begin{figure}
\centering
\includegraphics[width=\linewidth]{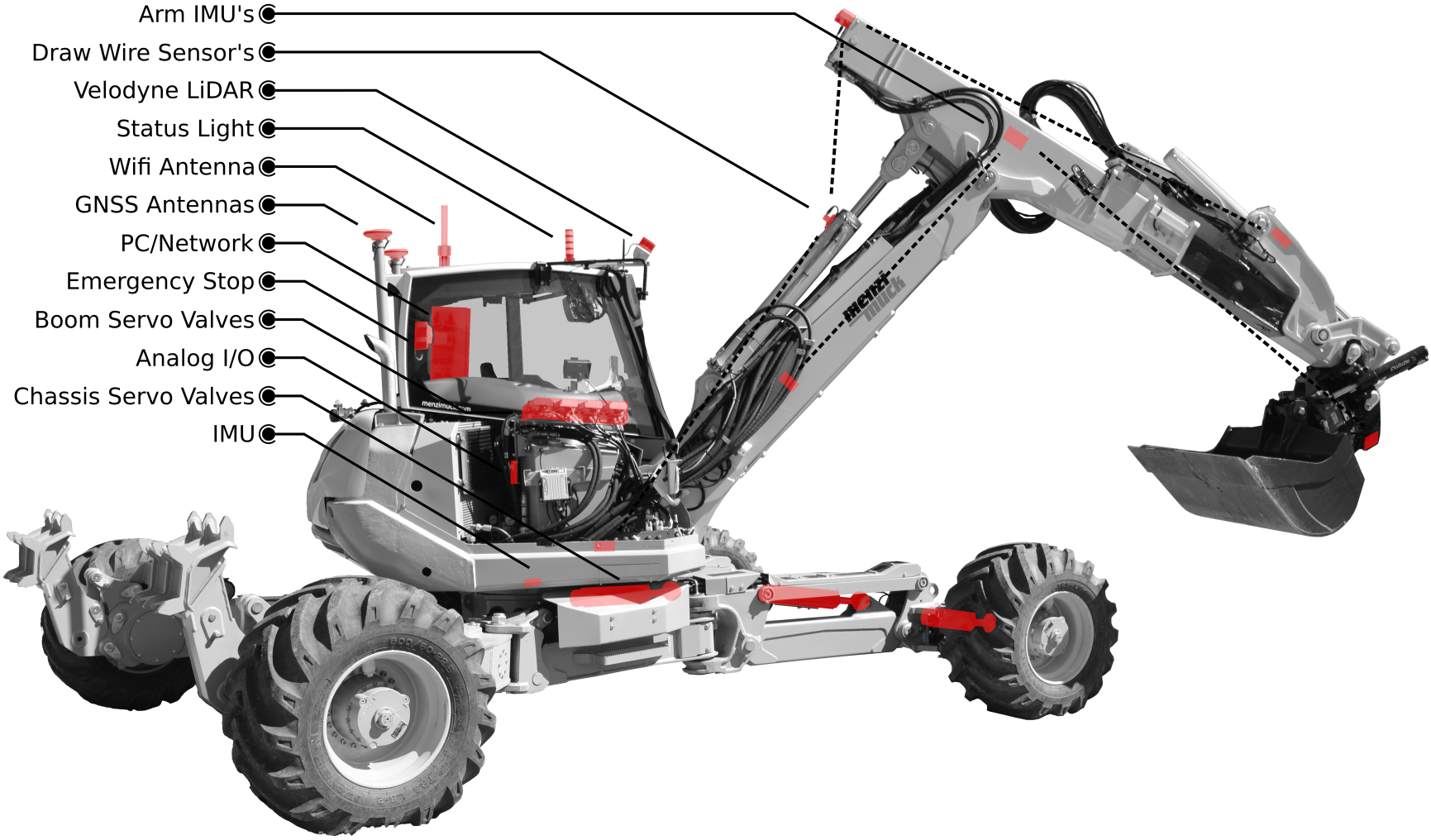}
\caption{The extensive hardware changes and main components added to a standard Menzi Muck M545 are highlighted.}
\label{fig:m545adaptations}
\end{figure}

HEAP is based on a 12 ton commercially available Menzi Muck M545 walking excavator, which has been highly customized with numerous adaptations and additions shown in Figure \ref{fig:m545adaptations}. Additionally, Table \ref{tab:joint_actuation} summarizes the available actuation principles and feedback measurements per joint.

\newcommand{\yes}{\textcolor{green!80!black}{\ding{51}}}
\newcommand{\no}{\textcolor{red}{\ding{55}}}
\newcommand{\yesno}{\textcolor{orange}{\ding{51}}}
\begin{table}
\setlength{\tabcolsep}{0.7\tabcolsep}
\centering
\caption{\label{tab:joint_actuation} HEAP's different joints have different capabilities. This table provides an overview of the possible measurements and realized actuation principles per joint. \yes\hspace{0.05cm}/\no\hspace{0.05cm} marks available/unavailable feedback or actuation. \yesno\hspace{0.05cm} indicates for the wheels that the velocity cannot be measured directly, but instead, it can be estimated from the chassis' linear velocity.}
    \begin{tabular}{ l c c c c c c c c c c c c }
        & \rotatebox[origin=l]{90}{Wheels} &  \rotatebox[origin=l]{90}{Steering} &  \rotatebox[origin=l]{90}{Ab-/Adduction} & \rotatebox[origin=l]{90}{Flexion/Extension} & \rotatebox[origin=l]{90}{Cabin Turn} & \rotatebox[origin=l]{90}{Boom} & \rotatebox[origin=l]{90}{Dipper} & \rotatebox[origin=l]{90}{Telescope} & \rotatebox[origin=l]{90}{Bucket Pitch} & \rotatebox[origin=l]{90}{Bucket Roll} & \rotatebox[origin=l]{90}{Bucket Yaw} \\ 
        \cmidrule(lr){2-5} \cmidrule(lr){6-12}
        &  \multicolumn{4}{c}{Chassis} & \multicolumn{7}{c}{Upper Machine} \\ \midrule
        \textbf{Feedback} \\
        Velocity      & \yesno & \yes & \yes & \yes & \yes & \yes & \yes & \yes & \yes & \yes & \yes\\
        Position      & \no    & \yes & \yes & \yes & \yes & \yes & \yes & \yes & \yes & \yes & \yes\\
        Force/Torque  & \no    & \yes & \yes & \yes & \yes & \yes & \yes & \yes & \yes & \no  & \no\\
        \textbf{Actuation} \\
        Servo Valve   & \no    & \yes & \yes & \yes & \no  & \yes & \yes & \yes & \yes & \no  & \no\\
        Pilot Stage   & \yes   & \no  & \no  & \no  & \yes & \yes & \yes & \yes & \yes & \no  & \no\\
        Act. Joystick & \no    & \no  & \no  & \no  & \yes & \yes & \yes & \no  & \yes & \no  & \no\\
        Rototilt R4   & \no    & \no  & \no  & \no  & \no  & \no  & \no  & \no  & \no  & \yes & \yes\\ \bottomrule
    \end{tabular}
\end{table}

\subsection{Sensing}
A Leica iCON iXE3 with two \ac{GNSS} antennas and a receiver is used for localization. \ac{RTK} corrections for the \ac{GNSS} signals are received over the internet from permanently installed base stations. SBG Ellipse2-A \ac{IMU}'s in both the cabin and the chassis complement these position sensors and allow for full 6-dof pose estimation of the cabin. The arm joint states are measured concurrently by draw wire encoders (Sick BCG05-C1QM0199) on the hydraulic cylinders and \ac{IMU}'s on the links. These two redundant systems offer different advantages and disadvantages. Draw wire encoders (or any other sensor measuring the cylinder position) are ideal for fast hydraulic control, because they measure the cylinder position accurately and without delay. However, joint play, inaccuracies in the conversion from piston position to joint position and a long kinematic chain from cabin to end-effector result in a worse pose estimate for the end-effector compared to the \ac{IMU}-based approach. \ac{IMU}'s measure the orientation of the links w.r.t. the world and are thus less prone to these errors. With proper dynamic compensation, they can even compete with cylinder based methods during dynamic motions. On HEAP, draw wires are used for control with servo valves and \ac{IMU}'s are utilized for precise end-effector pose estimation. Finally, two Velodyne Puck VLP-16 \ac{LiDAR}'s are used for perception. \ac{LiDAR}'s were chosen, because they outperform camera-based sensors in heavy dust environments and they provide more accurate and dense measurements compared to radar-based sensors \cite{Phillips2017lidarDust}.


\subsection{Software architecture}
The entire software framework runs on a computer (Intel i7-5820K, 6x3.60GHz, Ubuntu 18.04) installed in the cabin. The low-level controller acquiring the data from the sensors and controlling single actuators is triggered by the CAN driver at 100Hz, which in turn triggers the state estimator and high level controller. The high-level controller contains controllers and planners that generate joint commands for the arm and chassis. The time-critical communication between these software nodes happens over shared memory, while the information is additionally published using the \ac{ROS} for less time critical nodes, e.g. visualization tools. \ac{ROS} also allows control of the excavator, and displays any information gathered by the excavator on a remote computer over a network.


\subsection{Chassis Actuation}

To achieve active chassis balancing, all 14 stock hydraulic cylinders in the chassis were exchanged with new cylinders with an \ac{ICM} that allows for precise control of cylinder position and force \cite{Hutter2017Balance}. An \ac{ICM} includes an STM32F407 microprocessor running the cylinder control loops, pressure transducers and a servo valve. Revision 1 of these \ac{ICM}'s, as shown in \cite{Hutter2017Balance}, featured a valve similar to Moog G761. It is a servo valve with nozzle-flapper pilot stage technology that provides high dynamics, high resolution and low hysteresis. However, the nozzle was affected by the lower oil quality in a construction machine compared to hydraulics in other sectors. Oil contamination led to valve offsets. Furthermore, there is a pilot oil flow required. If all 14 actuators of the chassis are active, the high total pilot oil flow noticeably lowers the available power for the actual motion of the cylinders. Revision 2 was developed to circumvent these two major drawbacks. It has a direct drive valve (DDV), similar to a Moog D633, directly driven by a permanent magnet linear force motor with a high force level. It is strong enough to be unaffected by oil contaminants and does not need any pilot oil flow, but still preserves the positive high-performance characteristics of revision 1.

\subsection{Arm Actuation}

\begin{figure}
\centering
\includegraphics[width=\linewidth]{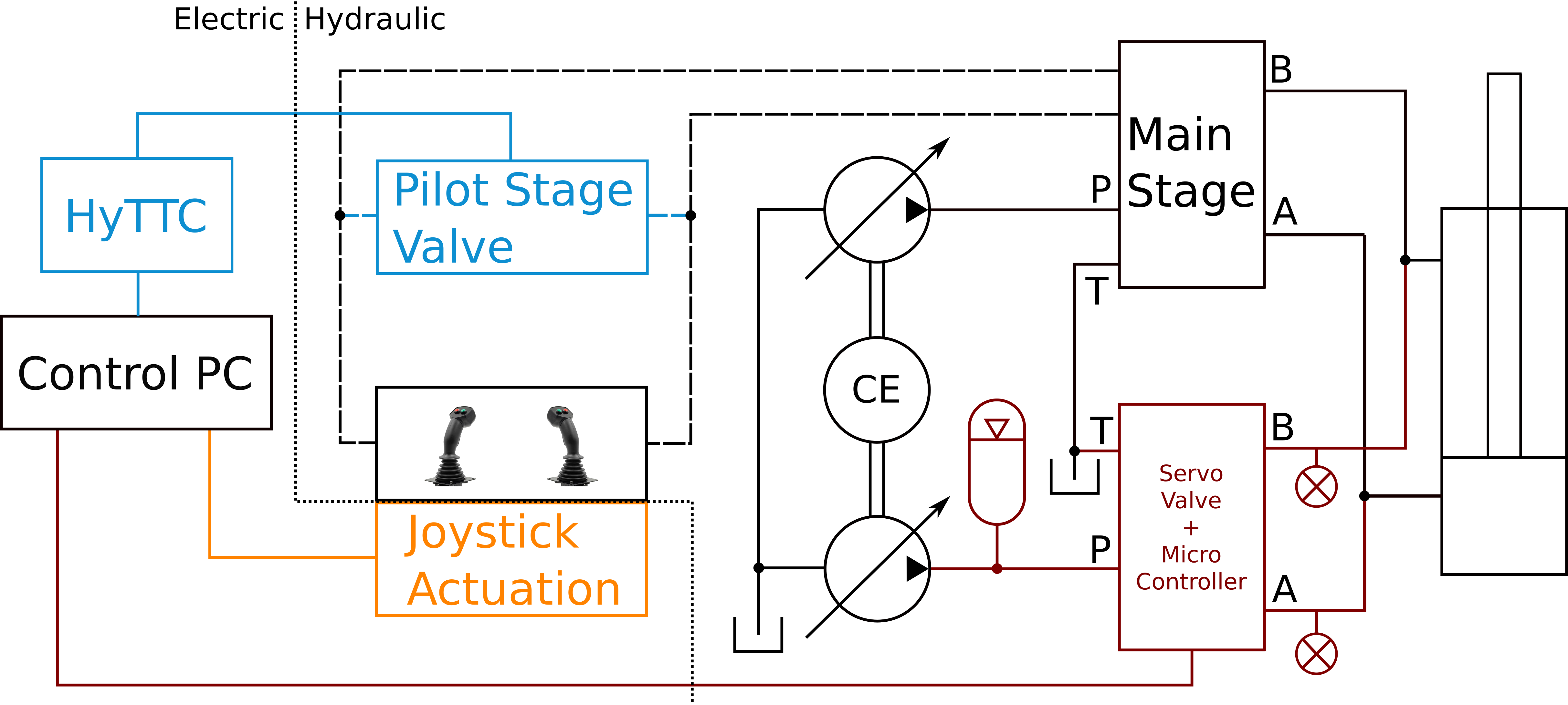}
\caption{The three different actuation types are joystick actuation (orange), electric pilot stage valves (blue) and Moog servo valves (red). The black lines and boxes represent the standard components already installed by the machine manufacturer. Note that a different pump is used for the servo valve as for the main stage valves. This second pump is normally used to supply large hydraulic attachments.}
\label{fig:hydraulics}
\end{figure}

The three most commonly found valve setups for automating the arm of an excavator are realized on HEAP. Firstly, a mechanism using two motors that physically move the installed hydraulic operator joysticks, specially developed for this work, was implemented. Secondly, electric pilot stage valves are installed in parallel to the joysticks, which reflects the most commonly used low-cost approach. Lastly, high-performance \ac{ICM}'s, as used in the chassis, are installed in the high pressure stage in parallel to the pilot stage driven main valves. Figure \ref{fig:hydraulics} shows the level of intrusion of these three different principles. Whereas the actuated joystick (orange) is minimally invasive and requires no tampering with the existing hydraulics (black), the pilot stage valves (blue) require the hydraulic low pressure circuit and the servo valves (red) in order for the high pressure circuit to be opened. The pre-installed main stage and the servo valves are supplied by two different pump systems. The main stage uses a traditional load sensing system that only raises the system pressure if necessary. The servo valves, on the other hand, are supplied by a constant pressure system with an accumulator. The advantage of the latter is that the full system pressure is immediately available without any delay, but with the downside of lower efficiency. 

The position and velocity of the pistons are measured by wire draw encoders, which are transmitting a velocity and position estimate at 100 Hz via CAN to the host computer. The cylinder force is estimated using pressure sensors integrated into the servo valve control modules. The estimated force is published at 100Hz via CAN for use with the pilot stage valves as well as the actuated joystick.

\subsubsection{Pilot stage operated main valves} \label{sec:actuatedJoystick}

Hawe PMZ proportional pressure reducing valves are installed in parallel to the joysticks in the pilot stage to control the standard main valves installed by the machine manufacturer. The input current of the valve is set by a TTControl HY-TTC 30 control unit connected to the CAN bus. The pilot stage is equipped with pressure sensors only for evaluation purposes and they are not used for control. The piston velocity 
is controlled using a feed-forward look-up table from desired piston velocities to valve currents and a PI-controller running at 100Hz to generate the reference current for the pilot valve. This is the most commonly used way of controlling heavy-duty hydraulic manipulators with valves that have a significant dead zone \cite{Nurmi2017}. The look-up table is identified over the entire range of applicable piston velocities with 13 value pairs. It improves the velocity control by providing a feed-forward term and compensating the large dead zone of the valve. This approach is facilitated by the load sensing system installed in a Menzi Muck excavator, as it allows use of the same look-up table for different load cases of the arm. The piston force is controlled with a PI-controller plus a dead zone compensation. 

\fboxsep=0mm
\fboxrule=2pt

\definecolor{joystickorange}{RGB}{255,131,0}

\begin{figure}
\centering
\subcaptionbox{Joystick actuation installed in the excavator.\label{fig:pic_joy}}{\includegraphics[width=0.48\linewidth]{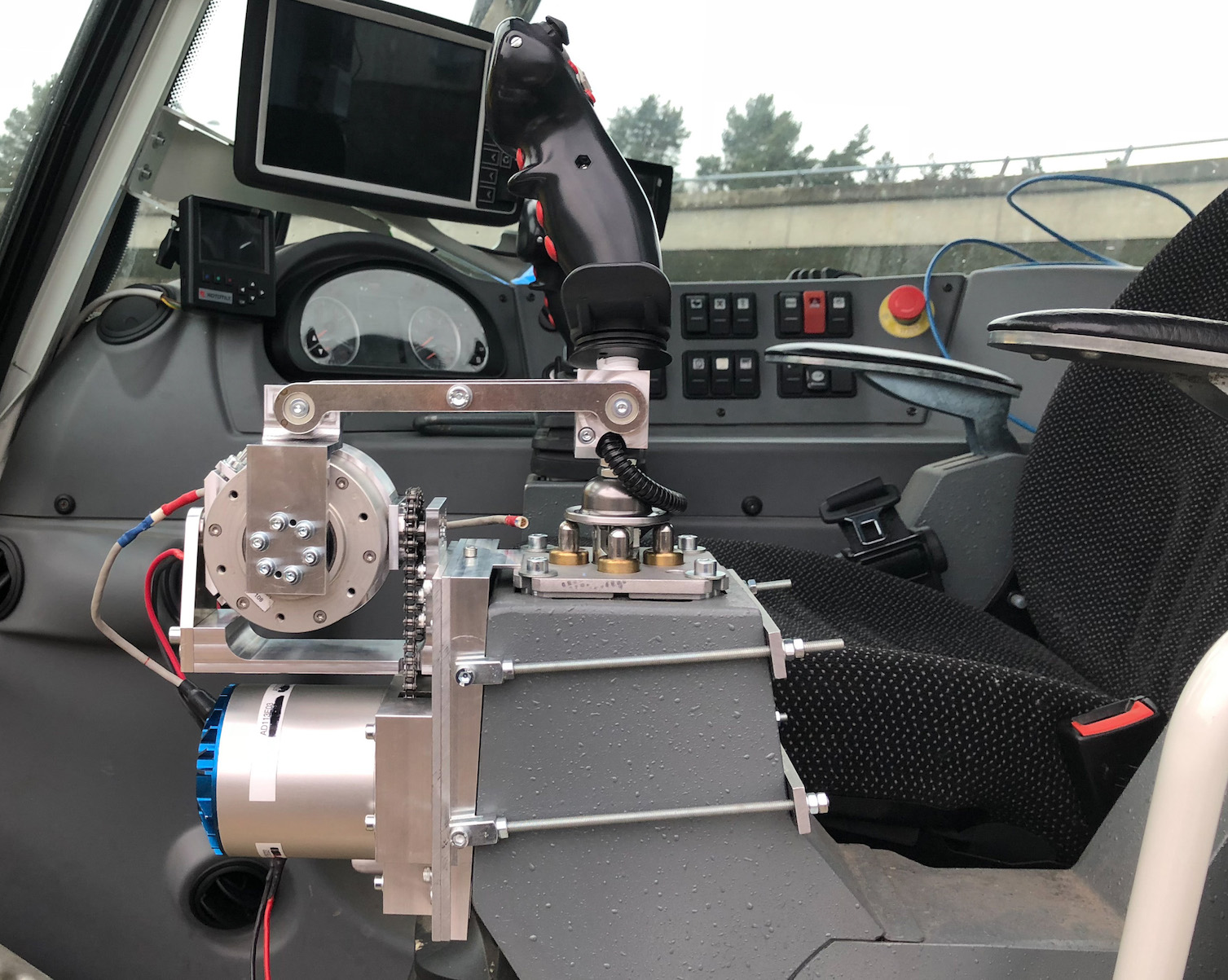}}
    \hfil
\subcaptionbox{CAD model of the joystick actuation with rotations axes marked as red dots.\label{fig:cad_joy}}{\includegraphics[width=0.48\linewidth]{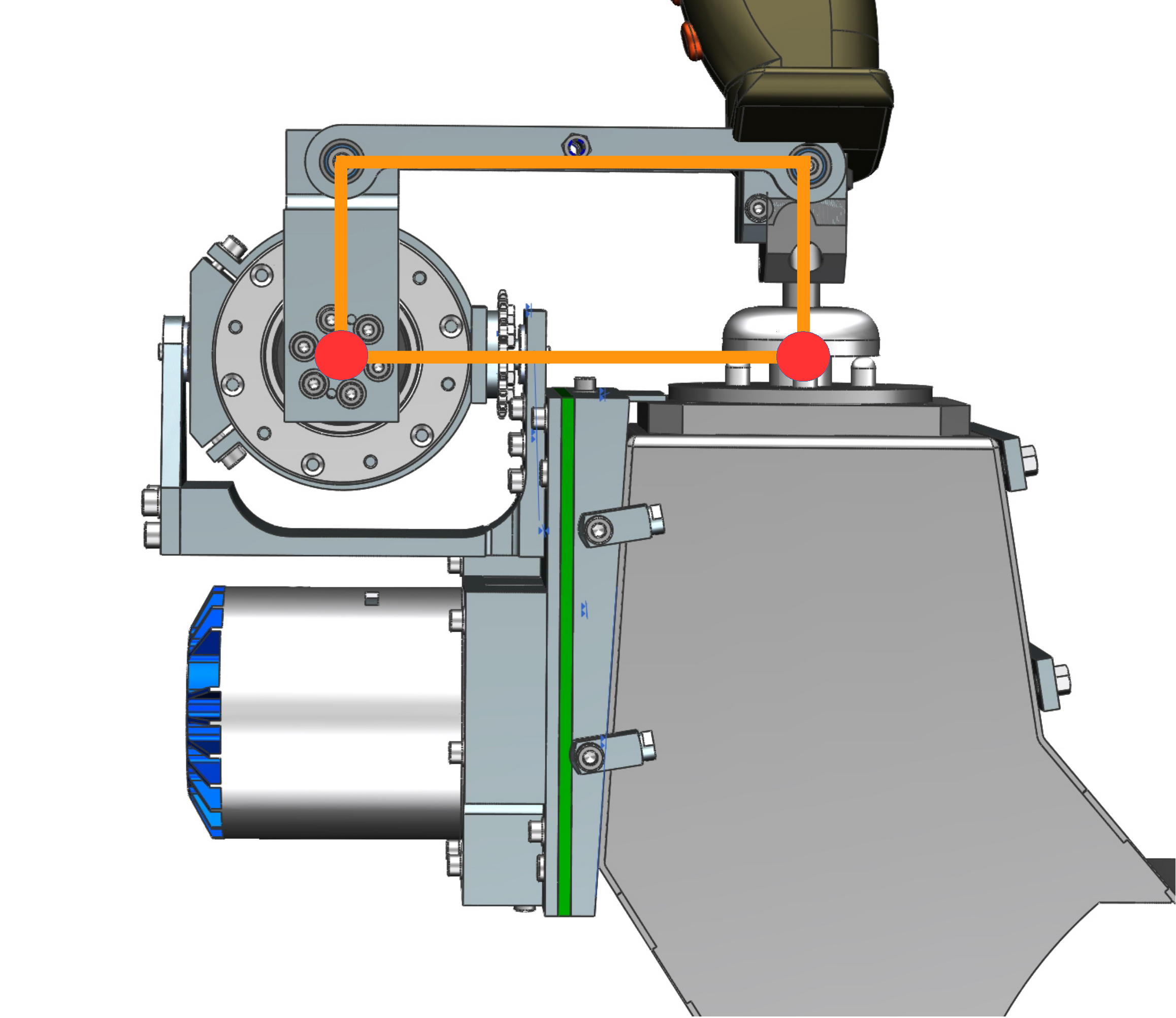}}
\caption{The actuated joystick mechanism can move the two axis of any regular excavator joystick.}
\label{fig:joy}
\end{figure}

The second option to utilize the installed main valves is to physically move the joysticks. Our actuated joystick mechanism uses two ANYdrive robotic joints \cite{Hutter2016patentAnydrive} to actuate the two main axes of one joystick. The ANYdrive's are position, velocity and torque controllable. The proposed solution is an enhancement of the work by Lee et al. \cite{Lee2015}. Lee's design results in a nonlinear mapping of motor angles to joystick angles and the required torques of the motors are unnecessarily high due to long lever arms. In our design, the rotation axes of the motors are lined up with the joystick's Cardan joint and positioned as close as possible. This results in the joystick angles being equivalent to the motor angles, removing the nonlinear mapping. The roll angle of the joystick is actuated by the bottom motor through a chain and the pitch angle with a parallel linkage illustrated with orange lines in Figure \ref{fig:cad_joy}. 

The torque transfer efficiency is defined as the ratio between the applied motor torque and the torque transmitted to the joystick. For our design, the torque transfer efficiency is $0.75$ in the worst-case configuration (maximum deflection on both axes). In Lee's design, the efficiency can go down to $0.3$ due to long lever arms. The rotation axis of the joystick and the mechanism is marked with red dots in Figure \ref{fig:cad_joy}. Two set screws allow for precise positioning of the two rotation axes w.r.t. each other by sliding the whole mechanism on the green plate shown in the CAD model. The joystick actuation was initially designed according to the maximum torque of 14.5Nm needed to move the joystick. It is calculated from the maximum force allowed on human-operated joysticks specified by the ISO standard for earthmoving machinery \cite{ISO} and the lever arm of the joystick. Thus, the  15Nm nominal torque of the ANYdrives would be a good fit. During testing, only a maximum of 1.73Nm was actually used to move the joystick. A second design iteration could, therefore, downscale the motors and make the unit more compact.

The piston velocity 
is controlled the same way as with the electric pilot stage. However, joystick angles are set instead of valve currents.

The high transparency of our actuated joystick in torque control allows for much more than simply an alternative to pilot stage or servo valves in autonomous control. The mechanism was specifically designed also to work with the operator sitting inside the machine and using it in the standard way. The actuated joystick can then enhance the operator's experience through haptic feedback, e.g., mapping the piston forces to the joystick axes or blocking the joystick if a certain depth is achieved with the shovel. 

\subsubsection{Servo Valve}

\fboxsep=0mm
\fboxrule=2pt

\definecolor{ruby}{RGB}{128,0,0}

\begin{figure}
\centering
\includegraphics[width=\linewidth]{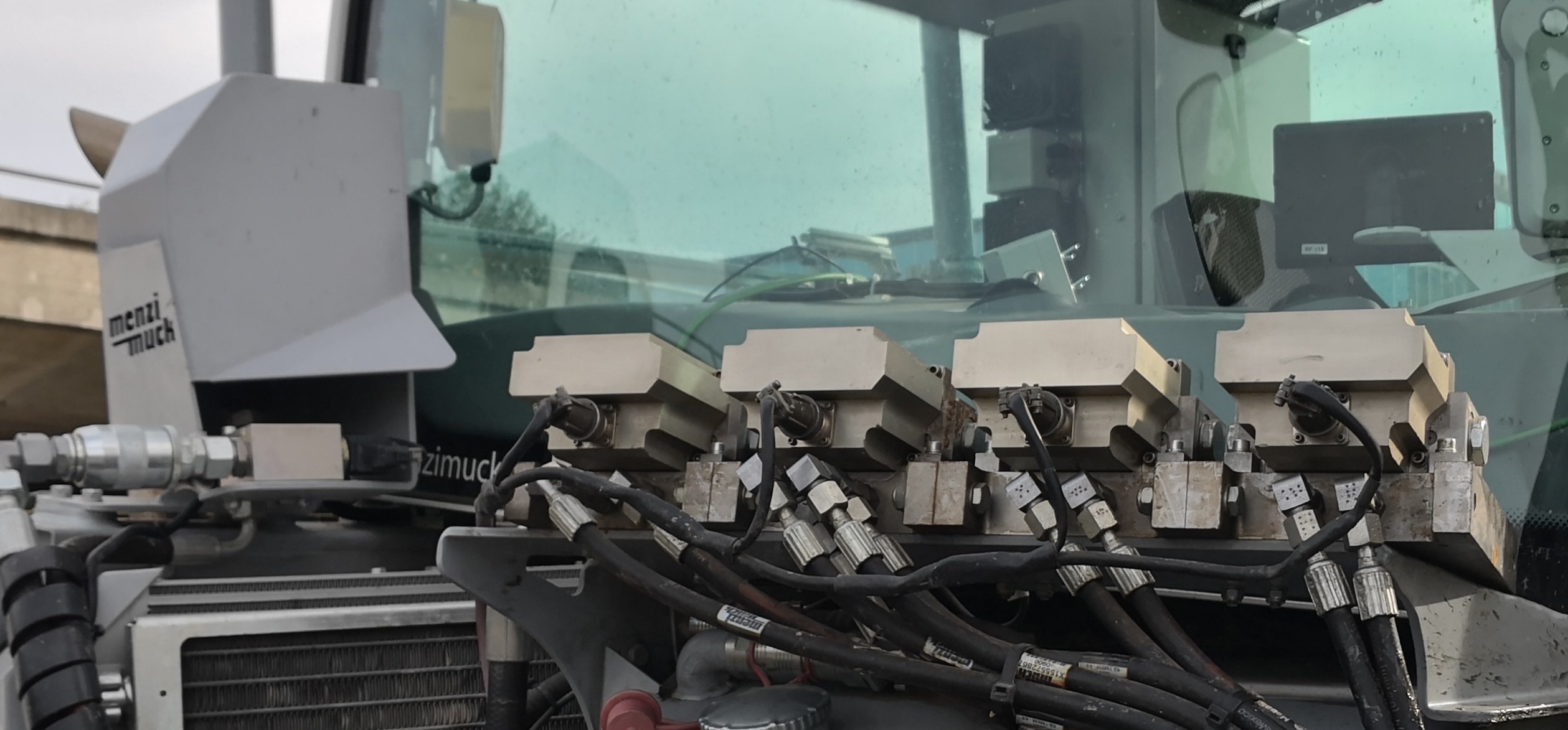}
\caption{The servo valves for boom, dipper, telescope and shovel joint together with the encapsulated accumulator in the supply line (on the left) are located next to the operator's cabin.}
\label{fig:arm_servo}
\end{figure}

\ac{ICM}'s are installed in parallel to the main valves in the high pressure circuit and consist of a high-performance servo valve with a current driver and a check valve for safety. The integrated ARM microprocessor can control piston position and velocity at 100Hz and force/pressure at 1kHz. The device communicates over CAN at 100Hz to the host computer. Piston position and velocity measurements are received via CAN from the draw wire encoder. The pressure sensors are directly connected to the microcontroller. Figure \ref{fig:arm_servo} shows the four servo valves for the boom, dipper, tele and bucket joint installed next to the cabin.

Servo valves will undoubtedly deliver a better performance than the pilot stage driven main valves in all aspects. However, servo valves can cause higher pressure drops that can stall the diesel engine. A hydraulic accumulator is installed on the supply line of the servo valves to support the pump in case of short high volume demands. It is depicted in Figure \ref{fig:arm_servo} next to the servo valves. As these valves were initially designed for smaller leg cylinders, they only allow for slower motion compared to the pilot stage driven main valves. The servo valves provide for simple modeling and control of the hydraulic system because there is no valve overlap. They are fast enough to compensate for various disturbances locally, achieving the required high precision end-effector control for accurate autonomous operation.

\subsubsection{Piston Velocity Control}\label{sec:pistonVel}

\definecolor{maroon}{RGB}{128, 0, 0}
\definecolor{orange2}{RGB}{255, 131, 0}
\definecolor{blue2}{RGB}{14, 143, 209}
\newcommand{\coloredcircle}[1]{\tikz\draw[#1,fill=#1] (0,0) circle (.75ex);}

\ifreview

\begin{figure}
  \centering
  \subcaptionbox{Step response.\label{fig:step_vel}}{\includegraphics[width=0.88\linewidth]{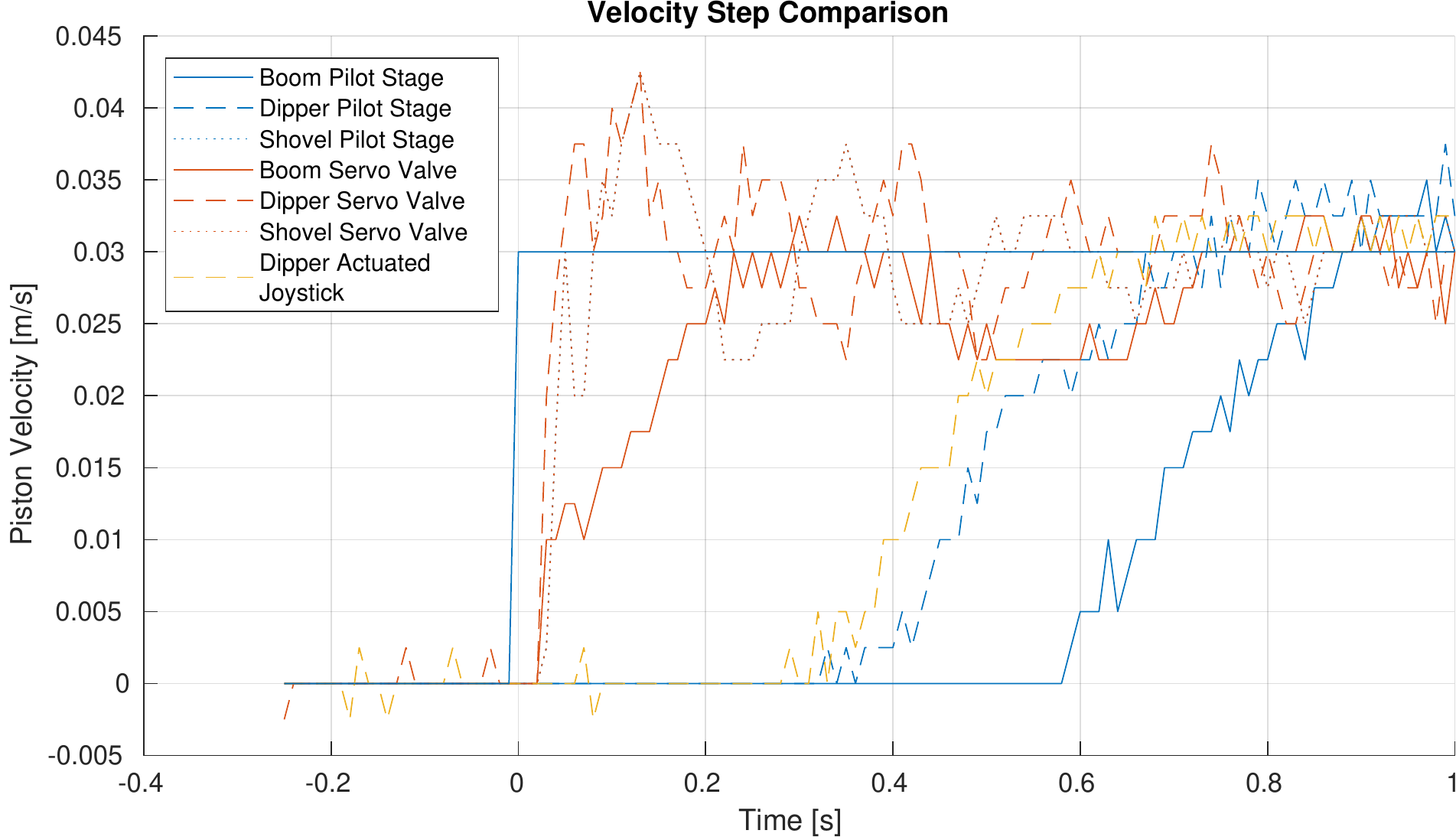}}
    \hfil
    \par\bigskip 
\subcaptionbox{Bode plot.\label{fig:bode_vel}}{\includegraphics[width=\linewidth]{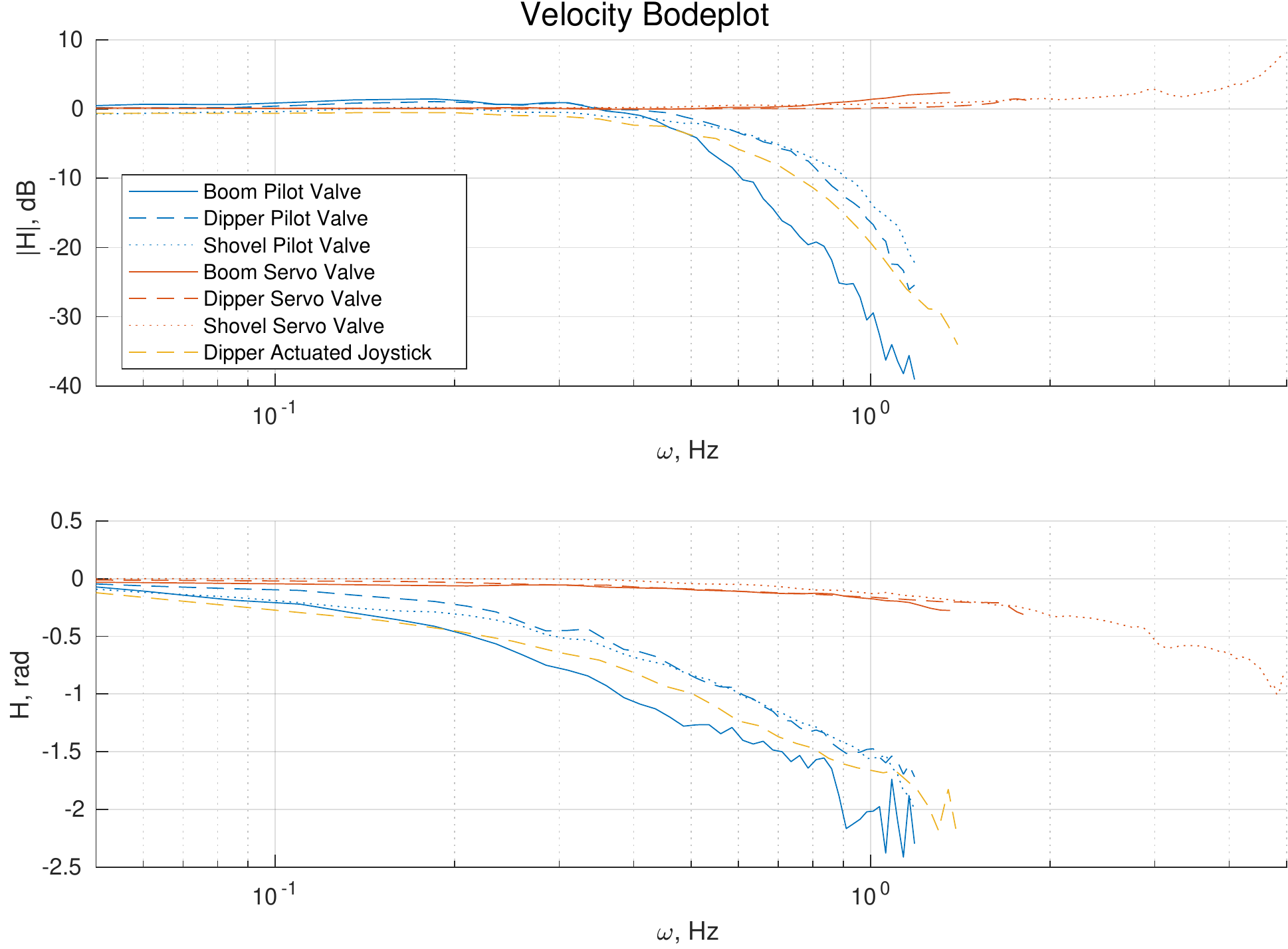}}
\caption{Velocity control comparison of servo valve (orange), pilot stage valve (blue) and actuated joystick (yellow) on the boom (solid line), dipper (dashed line) and shovel (dotted line) cylinder.}%
  \label{fig:step_bode_vel}
\end{figure}

\begin{table}
\renewcommand{\arraystretch}{1.5}
\centering
\captionof{table}[t]{\label{tab:vel_performance}The velocity control performance is evaluated by measuring the $t_{90}$ time and the cut-off frequency.}%
\vskip\abovecaptionskip
\begin{tabular}{lr@{}lr@{}l} \toprule
    \textbf{Velocity Control} & \multicolumn{2}{l}{\makecell[l]{$t_{90}$\\(along/against $g$)}} & \multicolumn{2}{l}{\makecell[l]{cut-off\\frequency}} \\ \midrule
    \coloredcircle{maroon} Servo Valves & $35$&$\si{\milli\second}$ & $>2.00$&$\si{Hz}$ \\
    \coloredcircle{blue2} Pilot Stage & $680$&$\si{\milli\second}$ / $850\si{\milli\second}$ & $0.50$&$\si{Hz}$ \\ 
    \coloredcircle{orange2} Actuated joystick & $600$&$\si{\milli\second}$ / -- & $0.45$&$\si{Hz}$\\\bottomrule
\end{tabular}
\end{table}

\else

\begin{figure}[t!]
\parbox{\linewidth}{\null
  \centering
  \subcaptionbox{Step response.\label{fig:step_vel}}{\includegraphics[width=\linewidth]{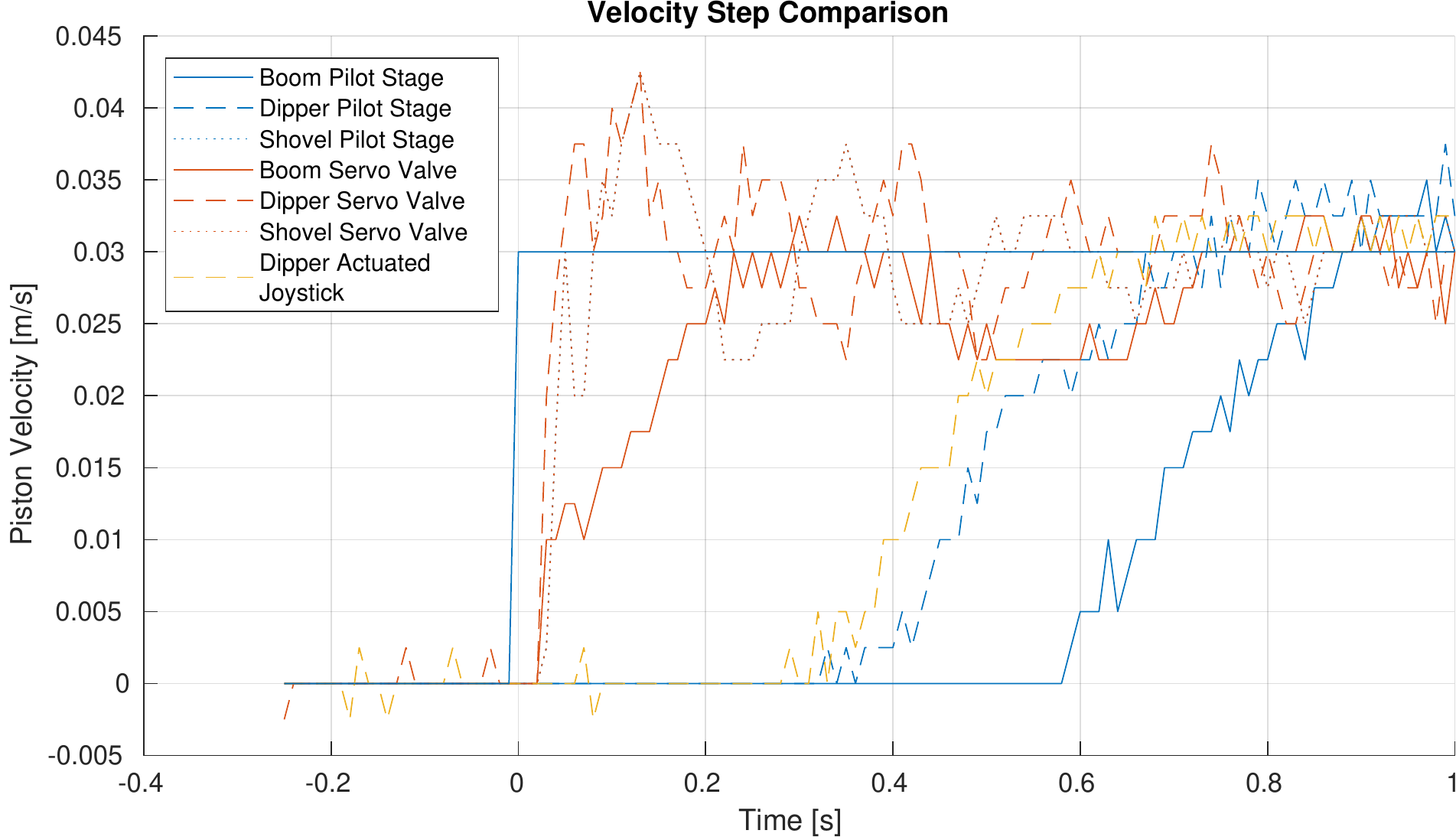}}
    \hfil
    \par\bigskip 
\subcaptionbox{Bode plot.\label{fig:bode_vel}}{\includegraphics[width=\linewidth]{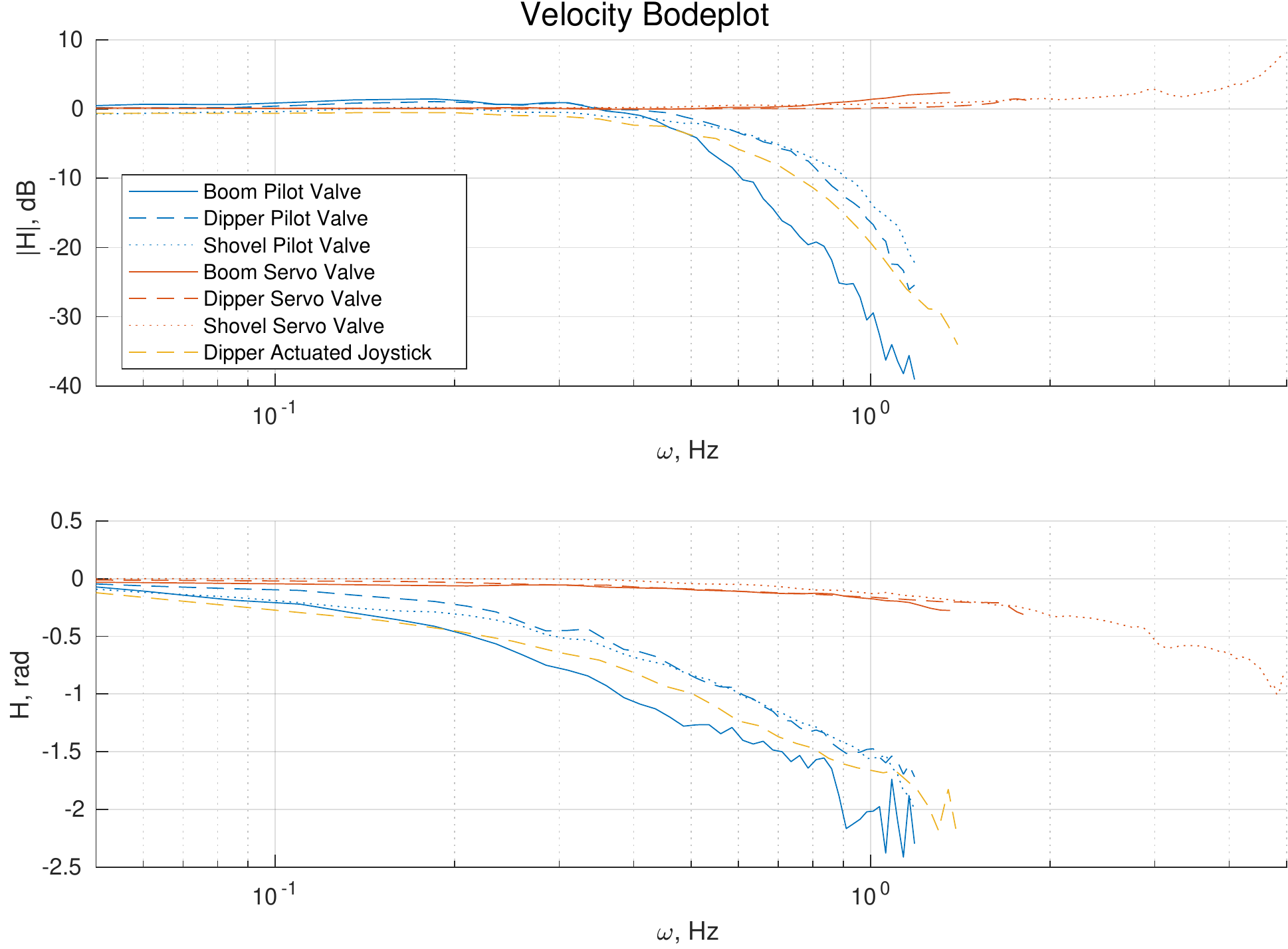}}
\caption{Velocity control comparison of servo valve (orange), pilot stage valve (blue) and actuated joystick (yellow) on the boom (solid line), dipper (dashed line) and shovel (dotted line) cylinder.}%
  \label{fig:step_bode_vel}
  \vspace{0.2cm}
}
\parbox{\linewidth}{\null
    \renewcommand{\arraystretch}{1.5}
    \centering
    \captionof{table}[t]{\label{tab:vel_performance}The velocity control performance is evaluated by measuring the $t_{90}$ time and the cut-off frequency.}%
    \vskip\abovecaptionskip
    \begin{tabular}{lr@{}lr@{}l} \toprule
        \textbf{Velocity Control} & \multicolumn{2}{l}{\makecell[l]{$t_{90}$\\(along/against $g$)}} & \multicolumn{2}{l}{\makecell[l]{cut-off\\frequency}} \\ \midrule
        \coloredcircle{maroon} Servo Valves & $35$&$\si{\milli\second}$ & $>2.00$&$\si{Hz}$ \\
        \coloredcircle{blue2} Pilot Stage & $680$&$\si{\milli\second}$ / $850\si{\milli\second}$ & $0.50$&$\si{Hz}$ \\ 
        \coloredcircle{orange2} Actuated joystick & $600$&$\si{\milli\second}$ / -- & $0.45$&$\si{Hz}$\\\bottomrule
    \end{tabular}
}
\end{figure}

\fi

Here, we compare the piston velocity control performance of the three different valve setups. Velocity control is mostly used for moving the arm in the air whereas force control is applied in case of large end-effector forces, e.g. when in ground contact. Thus, velocity control is evaluated for motions in the air. The input for the system identification of single actuators in velocity control mode was a chirp signal from $0.05\si{Hz}$ to $10\si{Hz}$ with an amplitude of $0.03\si[per-mode=fraction]{\meter\per\second}$. Step results are shown for step sizes of $0.03\si[per-mode=fraction]{\meter\per\second}$. The single actuator velocity control results are compared in Table \ref{tab:vel_performance}.

As expected, the servo valves have a quick step response to $90\%$ of the reference signal of $t_{90}=35\si{\milli\second}$, as shown in orange in Figure \ref{fig:step_vel}. Furthermore, it can be seen 
that the pilot stage valves have a large delay to the step reference caused by dead zones in the pilot valve as well as the main valve. The large difference between the dipper and boom/shovel pilot stage response time results from the fact that a positive step for the dipper goes with gravity, whereas the boom and shovel go against gravity. The driving force of gravity lets the joint move as soon as the main valve opens. Working against gravity, however, requires a certain system pressure, which is not immediately available due to the typical lag in such a hydraulic system. The response time of the dipper with gravity is $t_{90}=680\si{\milli\second}$ and of the boom and shovel against gravity $t_{90}=850\si{\milli\second}$. The actuated joystick achieves similar performance to the pilot stage valves for velocity control, as can be seen from the yellow curve in Figure \ref{fig:step_vel}. This supports the claim that most of the delay is caused by the main valves. Modern excavators have pressure sensors installed in the pilot stage (or are steer-by-wire), which allow for pressurizing the hydraulic system before the main valves open. However, this is not possible with this machine.

The bode plot in Figure \ref{fig:bode_vel} shows the gain and phase response of the three different velocity control loops. The identification for the servo valves had to be stopped before reaching the cut-off frequencies at $-3\si{\decibel}$ gain because the chirp frequency was close to the natural frequency of the excavator, which caused the entire machine to shake violently. It can only be concluded that the cut-off frequency lies well above $2\si{\hertz}$. Piston velocity control using the pilot stage valves achieved a cut-off frequency of $0.5\si{\hertz}$. The actuated joystick again produces a similar result to the pilot stage valves.

\subsubsection{Piston Force Control}\label{sec:pistonForce}
Here, force control is evaluated with the arm pressing on the ground. The input for the system identification of single actuators in force control mode was a chirp signal from $0.05\si{Hz}$ to $20\si{Hz}$ with an amplitude of $10\si{\kilo\newton}$. Step results are shown for step sizes of $20\si{\kilo\newton}$. The single actuator force control results are compared in Table \ref{tab:force_performance}.

\ifreview

\begin{figure}
  \centering
  \subcaptionbox{Step response.\label{fig:step_force}}{\includegraphics[width=0.94\linewidth]{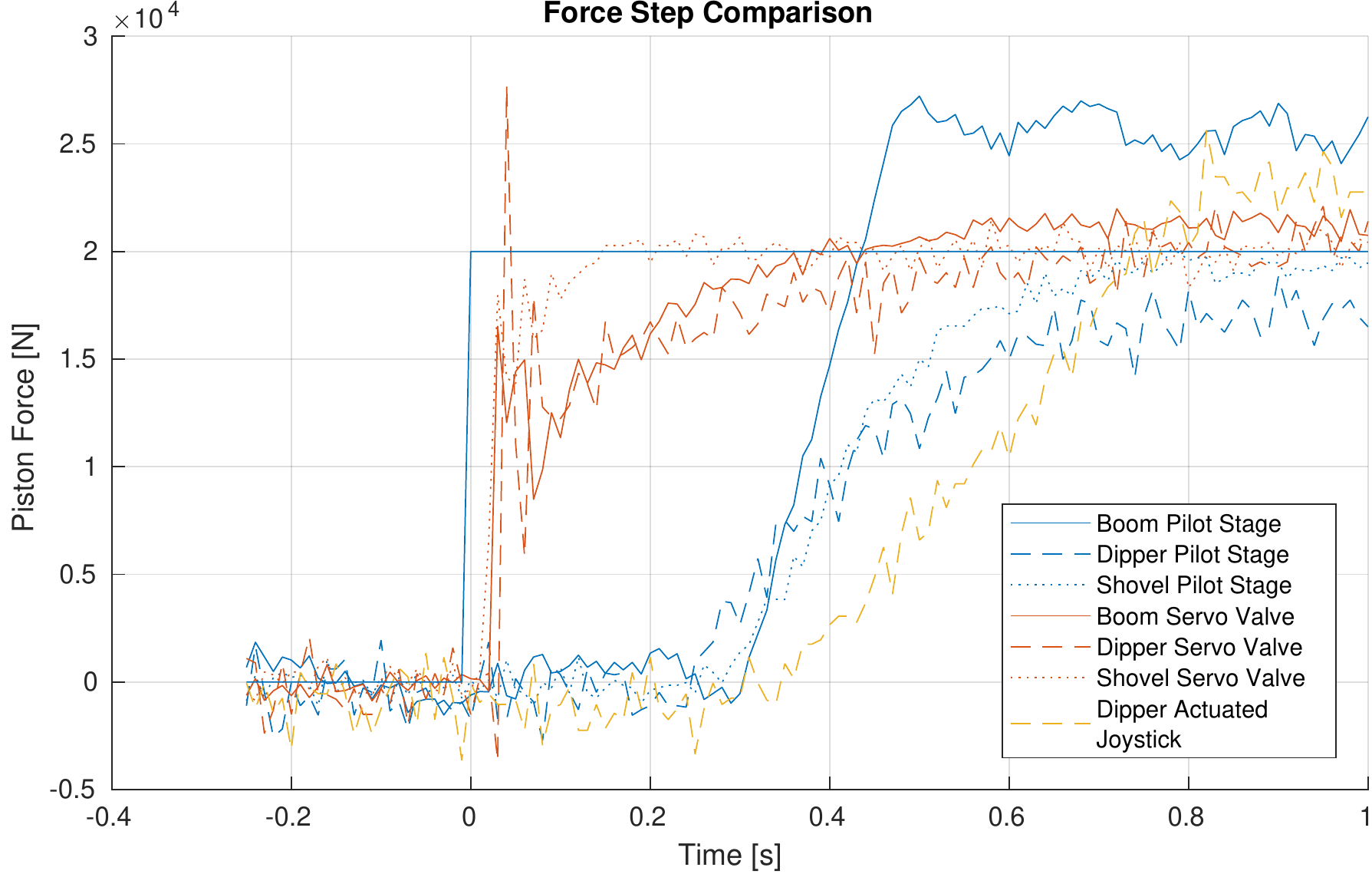}}
    \hfil
    \par\bigskip 
  \subcaptionbox{Bode plot.\label{fig:bode_force}}{\includegraphics[width=0.94\linewidth]{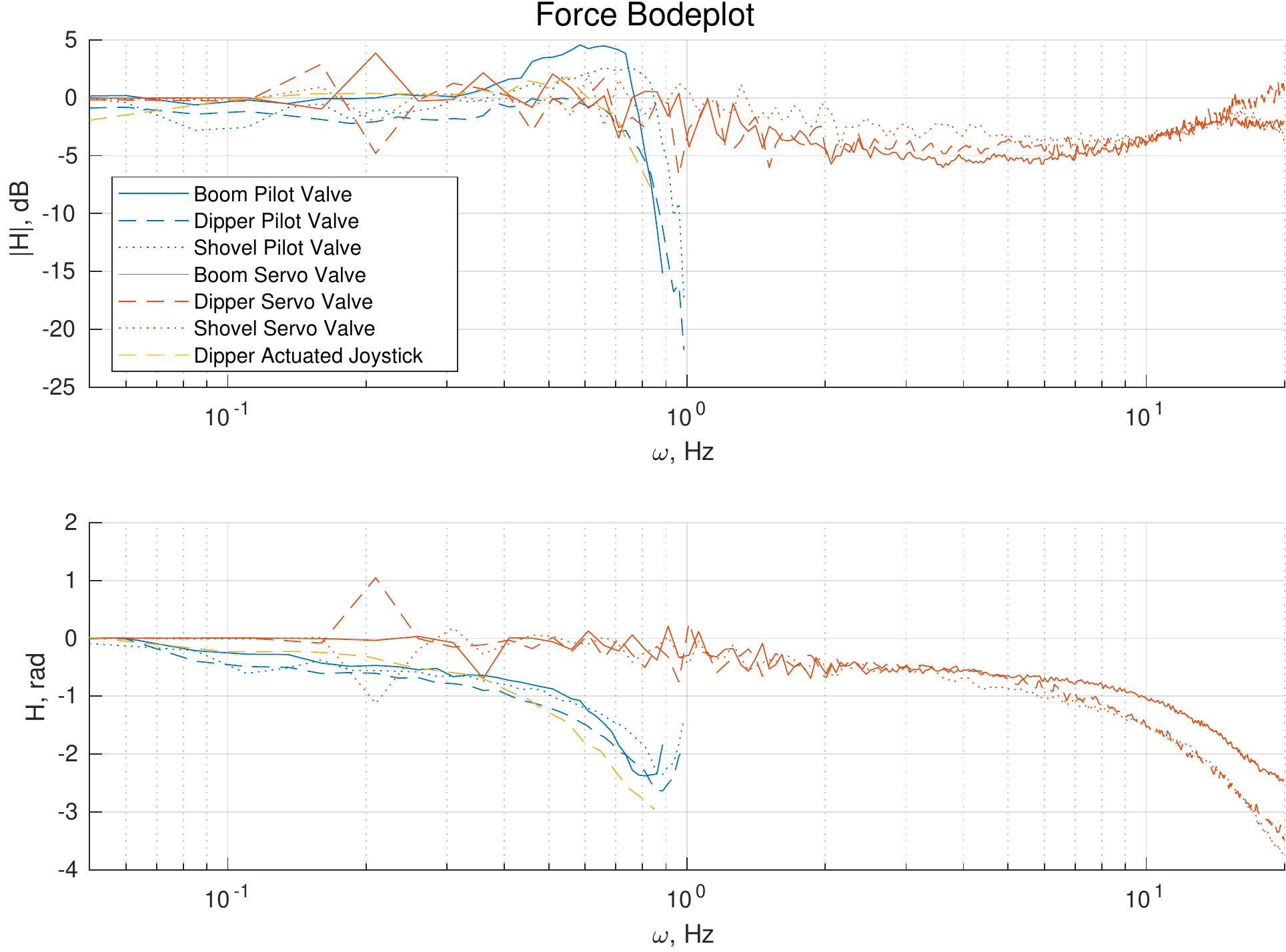}}
  \caption{Force control comparison of servo valve (orange), pilot stage valve (blue) and actuated joystick (yellow) on the boom (solid line), dipper (dashed line) and shovel (dotted line) cylinder.\label{fig:step_bode_force}}
\end{figure}

\begin{table}
    \renewcommand{\arraystretch}{1.5}
    \centering
    \captionof{table}[t]{\label{tab:force_performance} The force control performance is evaluated by measuring the $t_{90}$ time and the cut-off frequency.}
    \vskip\abovecaptionskip
    \begin{tabular}{llr@{}llr@{}l} \toprule
        \textbf{Force Control} && \multicolumn{2}{l}{$t_{90}$} && \multicolumn{2}{l}{cut-off frequency} \\ \midrule
        \coloredcircle{maroon} Servo Valves && $40$&$\si{ms}$ && $>20.00$&$\si{Hz}$\\
        \coloredcircle{blue2} Pilot Stage && $650$&$\si{ms}$ && $0.85$&$\si{Hz}$\\
        \coloredcircle{orange2} Actuated Joystick && $720$&$\si{ms}$ && $0.73$&$\si{Hz}$\\
        \bottomrule
    \end{tabular}
\end{table}

\else

\begin{figure}[t!]
\parbox{\linewidth}{\null
  \centering
  \subcaptionbox{Step response.\label{fig:step_force}}{\includegraphics[width=0.94\linewidth]{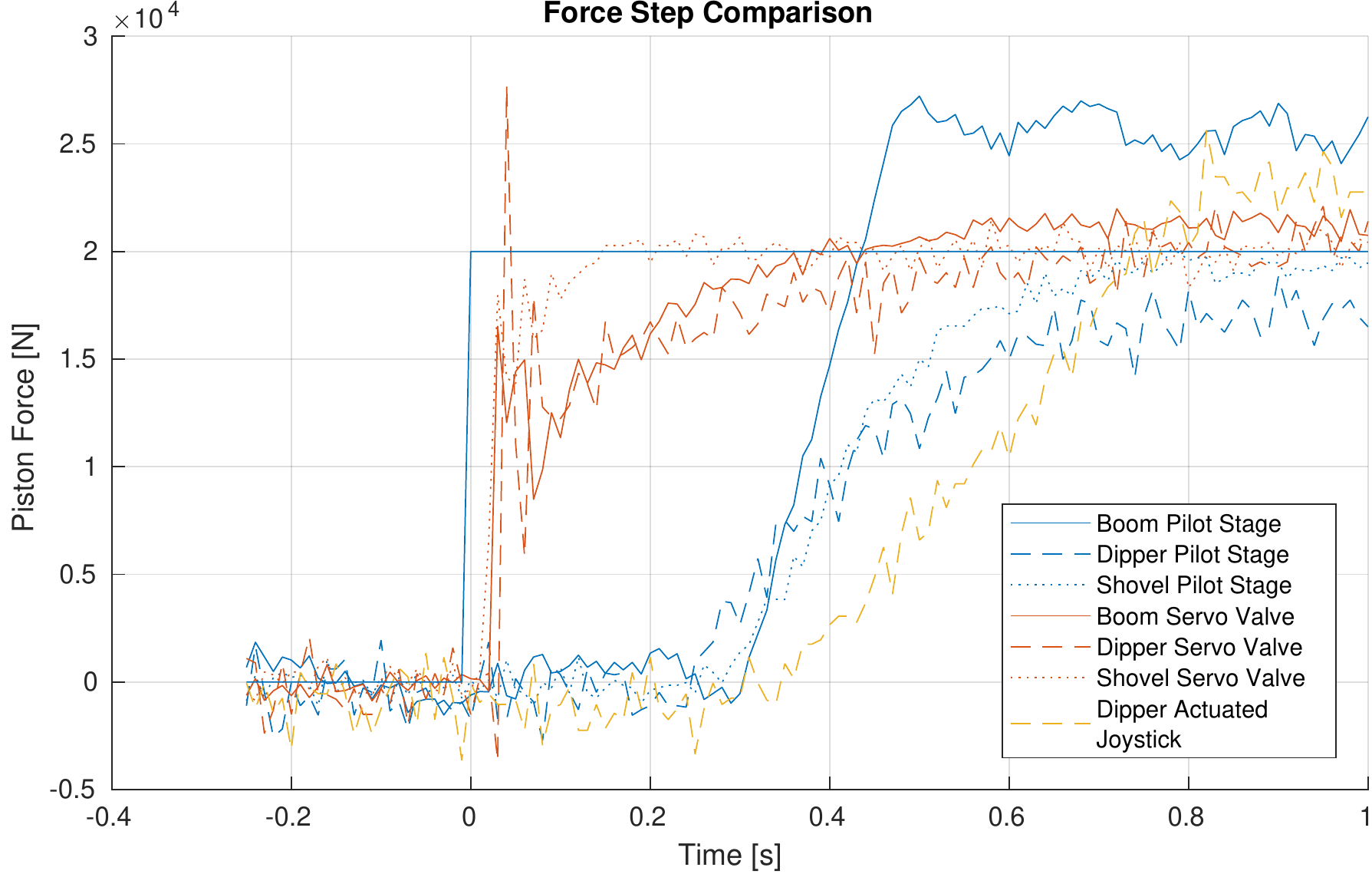}}
    \hfil
    \par\bigskip 
  \subcaptionbox{Bode plot.\label{fig:bode_force}}{\includegraphics[width=0.94\linewidth]{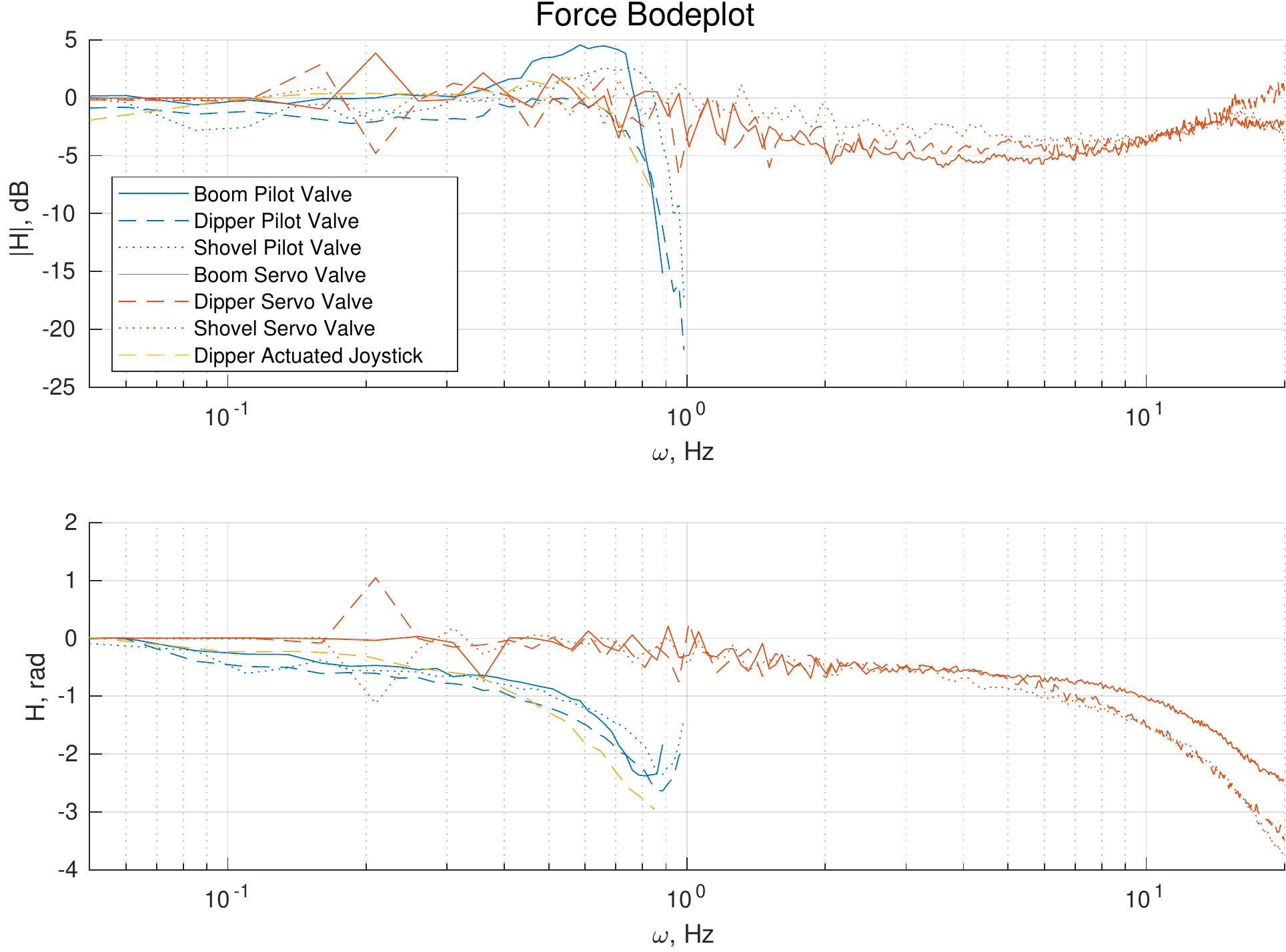}}
  \caption{Force control comparison of servo valve (orange), pilot stage valve (blue) and actuated joystick (yellow) on the boom (solid line), dipper (dashed line) and shovel (dotted line) cylinder.\label{fig:step_bode_force}}
  \vspace{0.2cm}
}
\parbox{\linewidth}{\null
    \renewcommand{\arraystretch}{1.5}
    \centering
    \captionof{table}[t]{\label{tab:force_performance} The force control performance is evaluated by measuring the $t_{90}$ time and the cut-off frequency.}
    \vskip\abovecaptionskip
    \begin{tabular}{llr@{}llr@{}l} \toprule
        \textbf{Force Control} && \multicolumn{2}{l}{$t_{90}$} && \multicolumn{2}{l}{cut-off frequency} \\ \midrule
        \coloredcircle{maroon} Servo Valves && $40$&$\si{ms}$ && $>20.00$&$\si{Hz}$\\
        \coloredcircle{blue2} Pilot Stage && $650$&$\si{ms}$ && $0.85$&$\si{Hz}$\\
        \coloredcircle{orange2} Actuated Joystick && $720$&$\si{ms}$ && $0.73$&$\si{Hz}$\\
        \bottomrule
    \end{tabular}
}
\end{figure}

\fi

Here, the servo valve responds in $t_{90}=40\si{\milli\second}$ to a force step input. These experiments were conducted with the excavator's arm as fixed on the ground as possible to avoid joint motion when a force is applied. However this was not always successful, as seen in the solid and dashed orange line in Figure \ref{fig:step_force}. The measured force shows a dip right after the first peak due to joint motion caused by the applied force. The pilot stage again has a large delay and struggles to match the reference force. Force control requires the valve to work often around zero opening, which is hindered by the large dead zone. The pilot stage valves achieve a rise time of $t_{90}=0.65\si{\second}$. 
The actuated joystick, with a step response time of $t_{90}=0.72\si{\second}$, is again able to keep up with the pilot stage .

The bode plot in Figure \ref{fig:bode_force} shows the gain and phase response of the servo valves (orange), pilot stage valves (blue) and the actuated joystick (yellow) in force control. The identification for the servo valves could only be done up to $20\si{\hertz}$. Signals with a higher frequency cannot be sampled correctly anymore at $100\si{\hertz}$ and would adversely affect the system identification. Thus, it can only be concluded that the cut-off frequency of the servo valves in force control lies above $20\si{\hertz}$. The dip in the gain from $1\si{\hertz}$ to $15\si{\hertz}$ is explained by the joint moving due to the applied force. For low frequencies up to $1\si{\hertz}$, the controller can track the reference force despite the joint motion. For higher frequencies above $15\si{\hertz}$, the mechanical system dampens the joint motion and the force tracking improves again. The pilot stage valves response cuts off at about $0.85\si{\hertz}$, but with a significant phase shift also at low frequencies. The gain and phase response of the actuated joystick is again similar to the pilot stage valves with a cut-off frequency at $0.73\si{\hertz}$.

\section{State Estimation of a Wheeled-Legged Robot}\label{sec:state_estimation}

The state estimator of HEAP is implemented as a general estimator for wheeled-legged robots using the two-state implicit filter (TSIF) of Bloesch et al. \cite{Bloesch_2018_tsif}. TSIF is a purely residual-based filter and therefore facilitates modularity and allows for a broader spectrum of models to be incorporated in the estimation compared to a more conventional Extended Kalman Filter. 

The position measurement of the cabin turn joint connecting the upper machine to the chassis is realized with two inductive sensors that detect passing gear teeth. An additional inductive sensor corrects potential errors of this incremental measurement when passing zero. Due to the low number of teeth in the large internal ring gear (98 teeth in total), an angular resolution of only 0.92$^\circ$ can be achieved. Thus fusing the chassis \ac{IMU} with the \ac{GNSS} measurement of the cabin would lead to a high error. Instead, \ac{GNSS} measurements are fused with the cabin \ac{IMU} and the state will be estimated in the cabin frame $C$, where the sensors are rigidly attached to each other. The chassis and cabin \ac{IMU}'s are used to augment the cabin turn angle measurement in a complementary filter. The cabin turn angular velocity $\dot{\psi}$ is estimated as the difference of the angular velocities of the cabin \ac{IMU} ($C$ frame) and the chassis \ac{IMU} ($B$ frame) in joint direction $z$ as
\begin{equation}
    \dot{\psi}' = {\tilde{\omega}_{BCz}} = {\tilde{\omega}_{ICz}} - {\tilde{\omega}_{IBz}} .
\end{equation}
The low-quality turn angle measurement $\tilde{\psi}$ from the inductive gear teeth sensors is augmented with the measured turn angular velocity in a complementary filter with the weighting constant $\alpha$ to compute the turn angle estimate $\psi'$ as follows
\begin{equation}
    \psi' = \alpha \tilde{\psi} + (1-\alpha)(\psi + \Delta t \dot{\psi}).
\end{equation}

The full state of this wheeled-legged robot can be computed using four residuals, namely \ac{IMU} prediction, \ac{GNSS} update (or any other position/orientation source), rolling prediction and legged odometry update, from two \ac{GNSS} measurements, cabin \ac{IMU} and all joint sensors. The estimated state contains 

\begin{align*}
    &_{I}{\boldsymbol{r}}_{IC} \text{, } _{I}{\boldsymbol{r}}_{IC}' \in \mathbb{R}^3 & &\text{position of cabin frame $C$ relative} \\
    &                                           & &\text{to inertial frame $I$ in $I$} \\
    &\Phi_{CI} \text{, } \Phi_{CI}' \in SO(3) & &\text{orientation of $C$ relative to $I$} \\
    &_{I}{\boldsymbol{v}}_{IC} \in \mathbb{R}^3 & &\text{linear velocity of cabin} \\
    &_{C}\boldsymbol{b}_f \in \mathbb{R}^3 & &\text{bias vector of linear acceleration} \\
    &_{C}\boldsymbol{b}_\omega \in \mathbb{R}^3 & &\text{bias vector of angular velocity} \\
    &{_{I}{\boldsymbol{p}}_{i}'} \in \mathbb{R}^3 & &\text{wheel contact position in $I$} \\
    &                                             & &i \in \left \{ rf,\ lf,\ lh,\ rh\right \} 
\end{align*}

$_{I}{\boldsymbol{r}}_{IC}$ and $\Phi_{CI}$ constitute the pose of the system. The biases are used to model the drift in the \ac{IMU} and angular velocity is excluded from the state, since it is measured directly with the \ac{IMU}.

Measurement noise is discrete-time Gaussian distributed ${\bar{\boldsymbol{w}}} \sim \mathcal{N}(0,Q)$. $Q \in \mathbb{R}^{3\times3}$ are tuning parameters of the estimation algorithm, capturing the properties of the different sensors as well as serving as weighting factors between different residuals operating on the same states.

Throughout this text, left subscripts will denote the frame in which a vector is expressed. States with primes refer to current states, whereas past states are not primed. See \cite{Bloesch_2016_primer} for references on the used calculus of 3D orientations.
\subsection{\acs{IMU} Prediction} \label{sec:imu}
Using measurements from the cabin \ac{IMU} one can formulate a straight forward prediction of the states ${_{I}\boldsymbol{r}}_{IC}$, $\Phi_{CI}$, ${_{I}\boldsymbol{v}}_{IC}$, ${{_C}\boldsymbol{b}}_{f}$ and ${{_C}\boldsymbol{b}}_{\omega}$. 

The \ac{IMU} measures the angular velocity of the cabin frame $_{C}\boldsymbol{\omega}_{IC} \in \mathbb{R}^3$ and the linear acceleration $_{I}\boldsymbol{a}_{IC}$ which splits into the gravity vector $_{I}\boldsymbol{g}$ and the proper acceleration ${_{C}\boldsymbol{f}_{IC}}$. There is stationary white noise $\boldsymbol{w}_f$ and $\boldsymbol{w}_\omega$ on the \ac{IMU} measurements, denoted $\tilde{\boldsymbol{f}}_{IC}$ and $\tilde{\boldsymbol{\omega}}_{IC}$. Adding the biases from the filter state to the \ac{IMU} measurements and integrating it into the Euler forward discretized (time step $\Delta t$) kinematics yields these prediction equations:

\begin{align}
    {_{I}\boldsymbol{r}}_{IC}' &= {_{I}\boldsymbol{r}}_{IC} + \Delta t\left[ {_{I}\boldsymbol{v}}_{IC} +\frac{{{_I}\bar{\boldsymbol{w}}_v}}{\sqrt{\Delta t}} \right]\label{pospred} \\
    \begin{split}
            {_{I}\boldsymbol{v}}_{IC}' &= \\ &{_{I}\boldsymbol{v}}_{IC} + \Delta t\left[ \Phi_{CI}^{-1}\left({_{C}\tilde{\boldsymbol{f}}}_{IC} - {_{C}\boldsymbol{b}_{f}} - \frac{{_{C}\bar{\boldsymbol{w}}_f}}{\sqrt{\Delta t}}\right) +{_{I}\boldsymbol{g}} \right]
    \end{split}\\
    \Phi_{CI}' &= \Phi_{CI} \boxplus \left[ -\Delta t\left({_{C}\tilde{\boldsymbol{\omega}}}_{IC}-{_{C}\boldsymbol{b}_{\omega}}-\frac{{_{C}\bar{\boldsymbol{w}}_\omega}}{\sqrt{\Delta t}}\right)\right]\label{orpred}
\end{align}

The ${_F}w_j$ have been discretized using zero-order-hold equivalence as $\frac{{_{F}\bar{\boldsymbol{w}}_j}}{\sqrt{\Delta t}}$ for $F \in \left\{ I,\ C \right\}$, $j \in \left\{ v,\ f,\ \omega \right\}$. We artificially introduce noise in the position prediction Equation \ref{pospred} acting as a weighting factor to possibly weaken the influence of the \ac{IMU} on position estimates. 

Also part of the \ac{IMU} residual is the prediction of the \ac{IMU} biases. Modelling the biases as Brownian Motions in a standard engineering allows us to formulate the prediction equation as
\begin{equation}\label{bpred}
    {_C{\boldsymbol{b}}'_{i}} = {_C{\boldsymbol{b}}_{i}} + \sqrt{\Delta t} {{_C}\bar{\boldsymbol{w}}}_{b,i}\, i \in \left \{ f,\ \omega \right \}.
\end{equation}

Equations \ref{pospred} to \ref{bpred} make up the full prediction equations for the \ac{IMU} residual.

\subsection{\acs{GNSS} Update} \label{sec:gnss}
 The \ac{GNSS} sensors provide absolute position measurements that allow the cabin to be located within the inertial frame. The innovation equation is again a comparison of the measurement and the sensor location computed from the estimator state. The innovation equation for the two sensors $k=\{1,2\}$ is
\begin{equation} \label{gpsinn}
\boldsymbol{y}_{\xi_k} = {_{I}\tilde{\boldsymbol{\xi}_k}} - {_{I}\boldsymbol{r}'_{IC}} - \Phi_{CI}'^{-1}({_{C}\boldsymbol{l}_{C\xi_k}}) + {_{I}\bar{\boldsymbol{w}}_{\xi_k}}.
\end{equation}

The \ac{GNSS} position measurement in inertial coordinates is represented by ${_{I}\tilde{\boldsymbol{\xi}_k}}$. The second and third terms in equation \ref{gpsinn} are a homogeneous transformation from the \ac{GNSS} sensor location in the cabin frame $C$, which is denoted ${_{C}\boldsymbol{l}_{C\xi}}$, into the inertial frame.

\subsection{Rolling Prediction} \label{sec:roll}

To use the contact points of HEAP's four wheels as landmarks, this residual implements a prediction of those contact points using measurements of the wheel speeds together with a kinematic model of the system to get wheel forward directions. This allows basic odometry for each contact point separately. In continuous time the kinematics for each contact point ${_{I}{p}_{i}}$ write
\begin{align}\label{rollpred}
{_{I}\dot{\boldsymbol{p}}_{i}} = \Phi_{BI}^{-1}( \rho \dot{\varphi}_i {_{B}{\bar{\boldsymbol{\nu}}_i}} +{{_B}\boldsymbol{w}}_{p,i})
\end{align}
There is white noise ${{_B}\boldsymbol{w}}_{p,i}$ on the contact point velocity. The wheel radii are assumed to be identical and represented by $\rho \in \mathbb{R}$, while $\dot{\varphi}_i \in \mathbb{R}$ denotes the $i$-th rotational wheel speed. Together with the wheel forward directions ${_{B}{\bar{\boldsymbol{\nu}}_i}} \in \mathbb{R}^3$ they make up the velocity of the contact points.

The ${_{B}{\boldsymbol{\nu}_i}}$ are projected into a plane defined by a terrain normal vector $\boldsymbol{n} \in \mathbb{R}^3$ to obtain ${_{B}{\bar{\boldsymbol{\nu}}_i}}$ as
\begin{equation}\label{proj}
{_{B}{\bar{\boldsymbol{\nu}}_i}} = {_{B}{\boldsymbol{\nu}_i}} - {_{B}{\boldsymbol{\nu}_i}}^{T} \Phi_{BI}({_{I}\boldsymbol{n}})\ \Phi_{BI}({_{I}\boldsymbol{n}})
\end{equation}
The terrain normal is assumed to be known for this and can be extracted from \ac{LiDAR} maps.

An Euler forward discretization of equation \ref{rollpred} now yields prediction equations for the contact points
\begin{align}\label{contactpred}
{_{I}\boldsymbol{p}'_{i}} = {_{I}\boldsymbol{p}_{i}} + \Delta t \ \Phi_{BI}^{-1}\left( \rho\dot{\tilde{\varphi}}_i {_{B}{\bar{\boldsymbol{\nu}}_i}} +\frac{{_B}\bar{\boldsymbol{w}}_{p,i}}{\sqrt{\Delta t}}\right)
\end{align}
with $\dot{\tilde{\varphi}}_i$ representing a discrete-time measurement of the wheel speed $\dot{\varphi}_i$. Combining equation \ref{proj} with \ref{contactpred} results in one innovation equation of the rolling prediction residual per leg
\begin{multline}\label{rollinn}
\boldsymbol{y}_{p,i} = {_{I}\boldsymbol{p}_{i}} + \Delta t \ \Phi_{BI}^{-1}\left( \rho\dot{\tilde{\varphi}}_i {_{B}{\boldsymbol{\nu}_i}} +\frac{{_B}\bar{\boldsymbol{w}}_{p,i}}{\sqrt{\Delta t}}\right) \\ - \Delta t \rho\dot{\tilde{\varphi}}_i \ {{_I}\boldsymbol{n}} {_{B}{\boldsymbol{\nu}_i}}^{T} \ \Phi_{BI}({{_I}\boldsymbol{n}}) - {_{I}\boldsymbol{p}'_{i}} 
\end{multline}

\subsection{Legged Odometry Update} \label{sec:kin}

The contact point landmarks predicted by the rolling prediction are now used to locate the base within the map composed of the four contact points. The innovation equations for this residual are a straight forward comparison of the contact points kept in the filter state on the one hand, and the same contact points computed from measurements of the joint positions on the other. In this way, the residual updates both the pose and contact points simultaneously by minimizing the discrepancy between the current filter state and the measured configuration of the legs. The actual measurements are the leg joint positions $\tilde{\boldsymbol{\alpha}} \in \mathbb{R}^{12}$, which via the HEAP kinematic model can be used to calculate vectors ${_{B}\boldsymbol{s}_{i}(\tilde{\boldsymbol{\alpha}})} \in \mathbb{R}^{3}$ from the base to the contact points. 

Those vectors can also be calculated as $\Phi'_{BI}({_{I}\boldsymbol{p}'_i} - {_{I}\boldsymbol{r}'_{IB}})$, which results in the following innovation equation.
\begin{equation}
\boldsymbol{y}_{s,i} = {_{B}\boldsymbol{s}_{i}(\tilde{\boldsymbol{\alpha}})} - \Phi'_{BI}({_{I}\boldsymbol{p}'_i} - {_{I}\boldsymbol{r}'_{IB}}) - {_{B}\bar{\boldsymbol{w}}_{s,i}}
\end{equation}
In case a wheel is not in ground contact, e.g. due to a stepping motion, the respective residual can be "deactivated" by setting ${_{B}\bar{\boldsymbol{w}}_{s,i}}$ to a high value, which will ignore this contact point as a landmark.

To follow \cite{imulegkin}, the $Q_{s,i}$ of ${_{B}\bar{\boldsymbol{w}}_{s,i}}$ are calculated from the uncertainties in the joint position measurements as well as the kinematic model. Firstly, the ${_{B}\boldsymbol{s}_{i}(\tilde{\boldsymbol{\alpha}})}$ are modelled to be affected by additive gaussian noise with zero mean and covariances $R_{s,i}$. Secondly, there is uncertainty in the joint position measurements $\tilde{\boldsymbol{\alpha}}$ themselves, which makes it necessary to propagate the covariance on the joint measurements through the model. Since most joint positions on an excavator are actually converted from piston displacements to joint angles by a mapping $f_{\alpha\beta}: \mathbb{R}^{12} \rightarrow \mathbb{R}^{12}$ such that $\boldsymbol{\alpha} = f_{\alpha\beta}(\boldsymbol{\beta})$ for the piston positions $\boldsymbol{\beta} \in \mathbb{R}^{12}$, the $Q_{s,i}$ are obtained from
\begin{equation}
Q_{s,i} = R_{s,i} + J_{1} J_{2} R_{\beta} J_{2}^T J_{1}^T
\end{equation}
using the two Jacobians $J_{1} = \frac{\partial{_{B}\boldsymbol{s}_{i}}}{\partial \boldsymbol{\alpha}}$ and $J_{2} = \frac{\partial f_{\alpha\beta}}{\partial \boldsymbol{\beta}}$. $R_\beta$ corresponds to the actual covariance of additive zero-mean gaussian noise on the piston measurements.

\subsection{Implementation On A Walking Excavator} \label{sec:res_estimation}

\begin{figure}
    \centering
    \includegraphics[width=\linewidth]{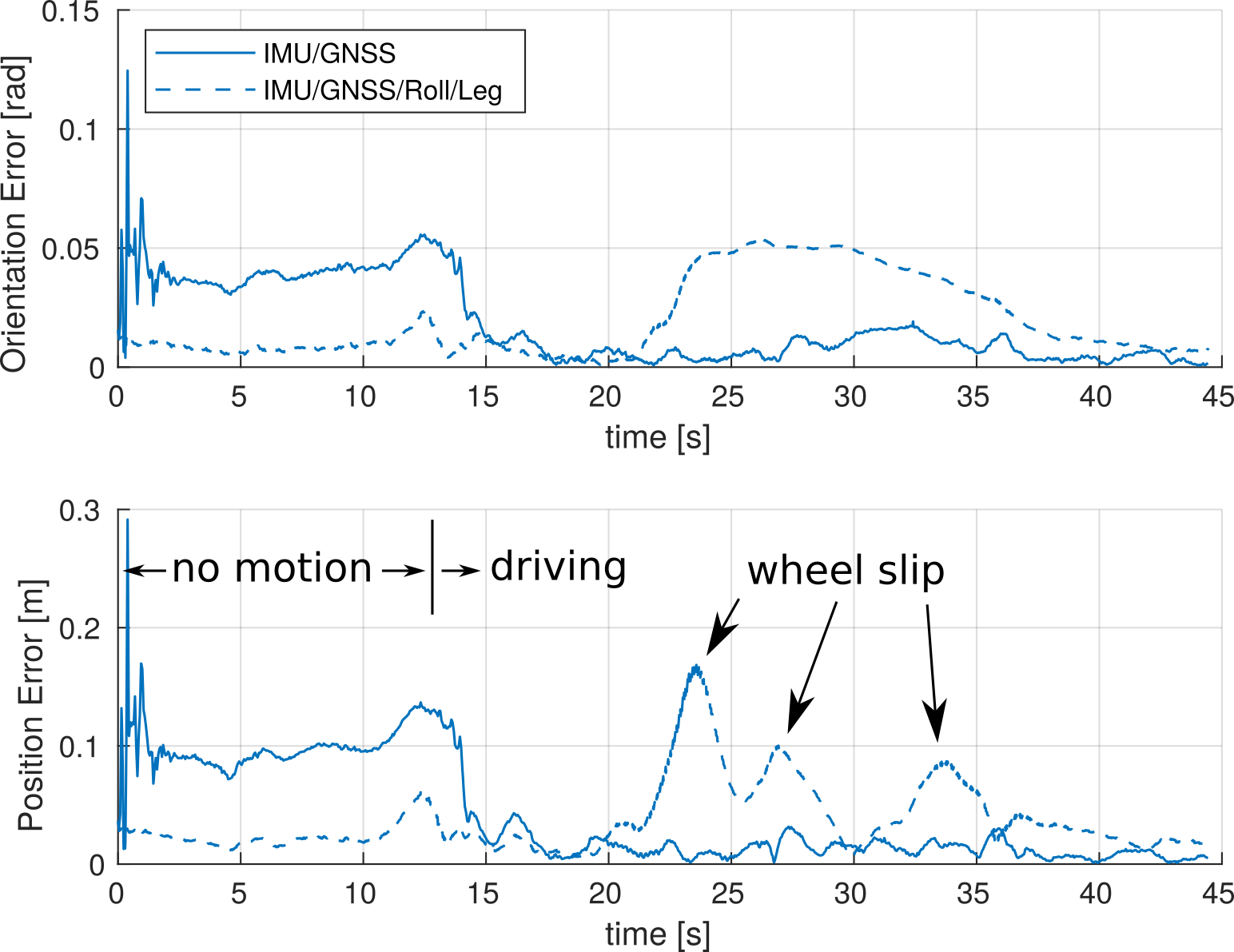}
    \caption{The sum of squared errors in orientation is depicted in the upper plot and for the position in the lower plot. The estimator is initialized with unknown biases at time 0s and the machine stays stationary until time 12s, then it starts to move the legs and drive. The \ac{IMU}/\ac{GNSS} approach (solid line) shows higher initial errors, but outperforms the \ac{IMU}/\ac{GNSS}/Roll/Leg approach (dashed line) once the machine starts to move.}
    \label{fig:est_error}
\end{figure}

The implementation of this estimator for HEAP was evaluated in a Gazebo simulation environment, which provides accurate ground truth on all estimated states. The individual sensors are simulated with realistic noise and drift parameters. First, \ac{IMU}, \ac{GNSS}, rolling and legged odometry residuals are combined in an attempt to feed as much information to the estimator as possible. Figure \ref{fig:est_error} shows the sum of squared errors in position and orientation of this setup in the dashed line. The estimator does not perform well during yawing (driving a tight corner) and acceleration and deceleration. This is caused by the rolling constraint that cannot account for wheel slip. Thus, we tested a second residual setup that consists only of \ac{IMU} and \ac{GNSS} residuals. Its performance is shown by the solid line in Figure \ref{fig:est_error}. A large error is shown at the beginning after the estimator is initialized and disappears after the robot starts to move. The cause is that \ac{IMU} biases cannot be estimated without motion in this setup. Once the biases have converged, the \ac{IMU}/\ac{GNSS} setup outperforms the full setup for aggressive motions.

Following this result, the \ac{IMU}/\ac{GNSS} setup was deployed on the system. This was motivated not only by the fact that rolling and kinematic constraints could not cope well with wheel slip in simulation, but also because on the real machine wheel speeds cannot be measured easily, yet in the hydraulic wheel hub motor and wheels tend to slip even more. Despite this drawback, the derivation of rolling and kinematic constraints is still a large contribution: our estimator is the first one to properly fuse leg kinematics, IMU measurements and some positioning source and can be applied to any wheeled-legged robot. In the case of our walking excavator, the rolling and kinematic residuals can also be used in the future for detecting wheel slip, which will become important for autonomous missions in more difficult terrain. Additionally, activating rolling and kinematic constraints whenever the GNSS reception is lost (essentially reducing it to dead reckoning), will prevent the estimator from drifting away when only integrating IMU measurements.

\section{Control} \label{sec:control}
\begin{figure}
    \centering
    \includegraphics[width=\linewidth]{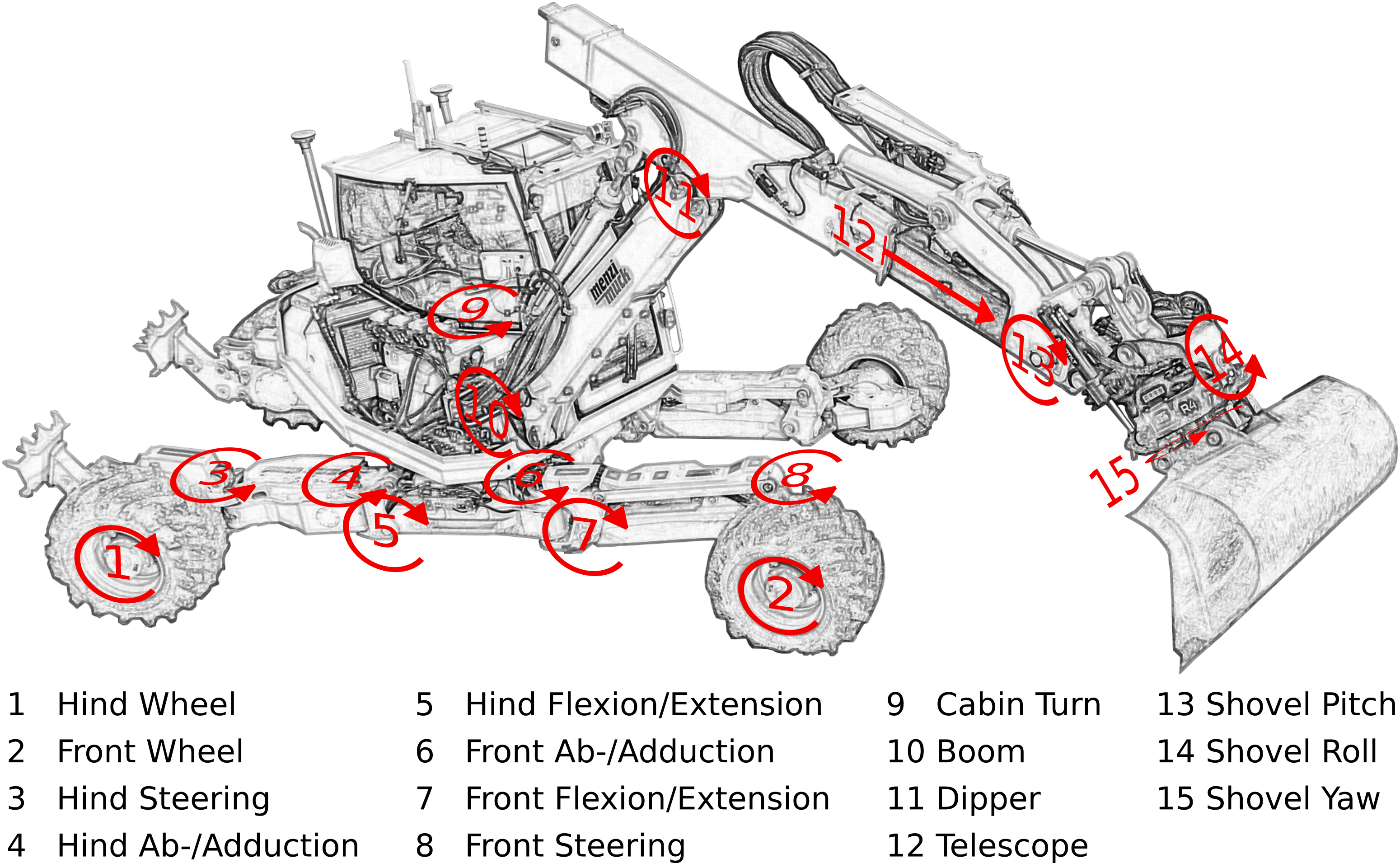}
    \caption{The 18 joints belonging to chassis are exemplarily indicated on the right hind and front leg with numbers 1 to 8, excluding the claw on the hind leg. Joints 9 through 15 belong to the arm.}
    \label{fig:joints}
\end{figure}

HEAP has a total of 31 degrees of freedom spread over five limbs (four legs and one arm). The joint axes and names are shown in Figure \ref{fig:joints}. Joints 1-8 belong to the right front and hind leg. The claws on the hind legs are two additional DOF's that are not used in autonomous mode yet. They are used by expert operators to avoid sliding on steep terrain in our teleoperation setup. Joints 9-15 belong to the arm. Cabin turn, boom, dipper, telescope and shovel pitch are standard joints for excavators, which are accompanied by two additional DOF's from a Rototilt R4.

\begin{figure}
    \centering
    \includegraphics[width=\linewidth]{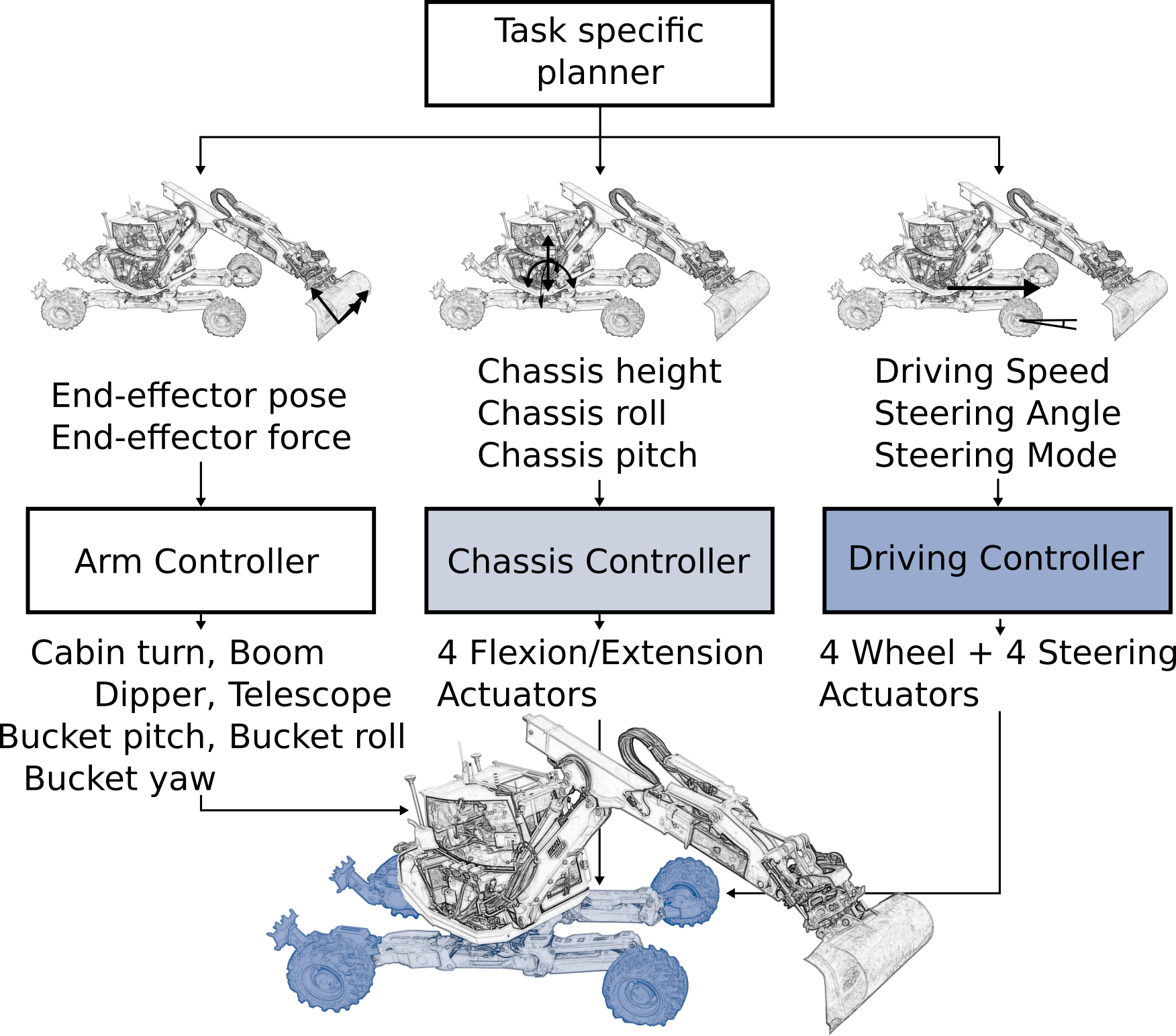}
    \caption{The machine is controlled through three individual controllers. First, an arm controller uses the seven arm joints of the upper machine to achieve an end-effector pose or force. Secondly, a chassis controller can adapt to legs to the ground and control height, roll and pitch of the machine's base with the four flexion/extension cylinders. Lastly, the driving controller utilizes four steering and wheel joints to control driving speed and direction.}
    \label{fig:controllers}
\end{figure}

Because of these many DOF's, the given topology and the high complexity of the machine, motivates us to split the control of the machine into three parts. Figure \ref{fig:controllers} illustrates the input of these three controllers and highlights the part of the machine and joints they use. First, an arm controller tracks end-effector poses and forces using the seven arm joints. Secondly, a chassis controller can adapt the legs to the ground and control height, roll and pitch of the base with the four flexion/extension joints. Lastly, a driving controller is used to compute the wheel speeds from the desired driving velocity and the angle of all four steering actuators from a desired steering angle and steering mode depending on the leg joint angles. We assume the existence of a task-specific planner, omitted in this article, that generates the inputs for these three controllers.

\subsection{Driving Controller}

The input of the driving controller is the desired driving velocity, steering angle and steering mode. The velocity of the wheels cannot be individually controlled, because the hydraulic wheel hub motors are connected in series. Instead, a velocity input is applied to all four wheels simultaneously. The wheels are not equipped with encoders, meaning that individual wheel speeds cannot be measured and the driving velocity is thus controlled by feeding back the linear velocity of the base from the state estimator.

\begin{figure}
    \centering
    \includegraphics[width=\linewidth]{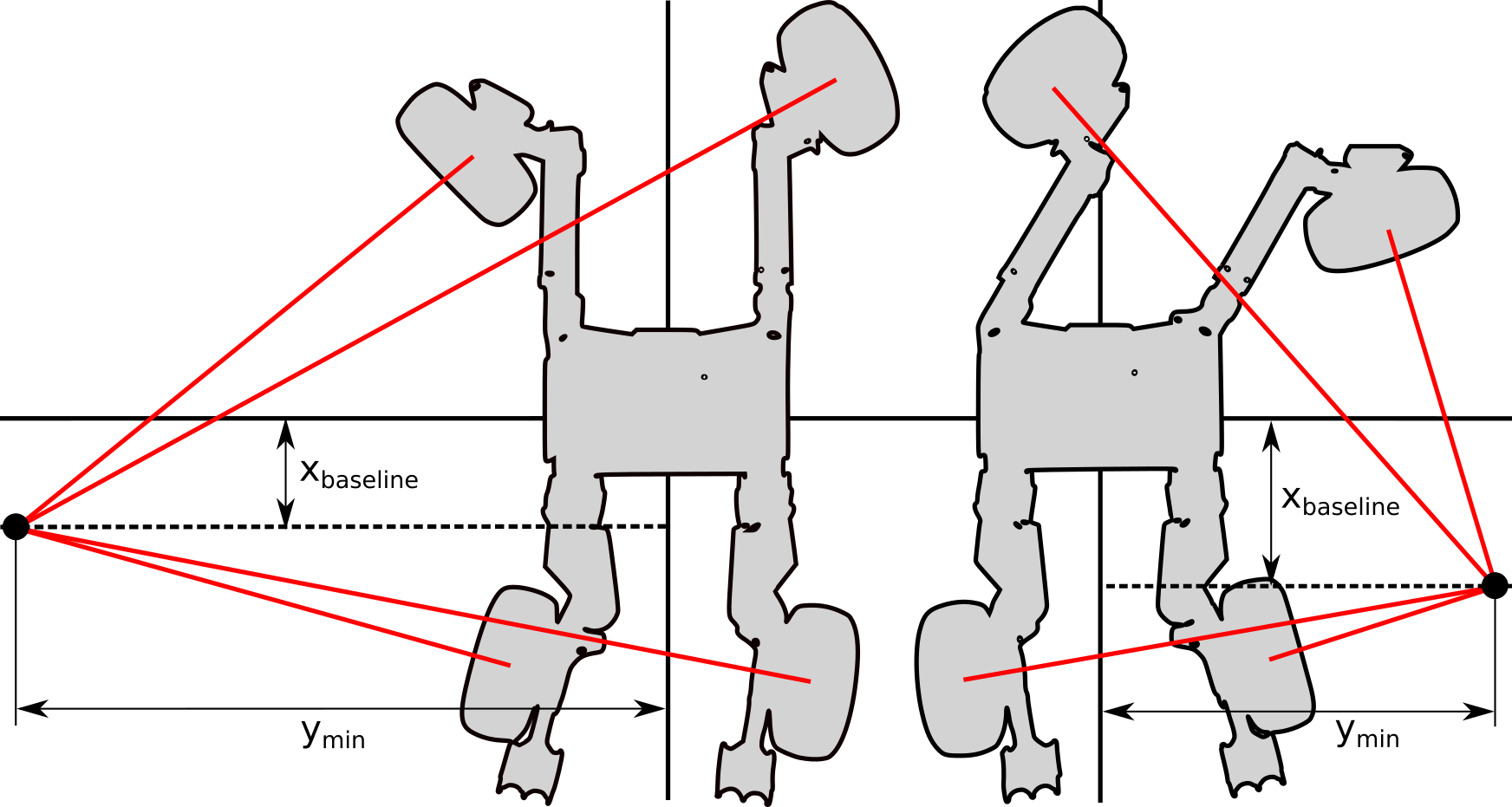}
    \caption{The minimum turning radius with four wheel steering depends on the hip ad-/abduction angles. The left configuration shows zero hip ad-/abduction, whereas the left configuration uses the hip ad-/abduction for a tighter turn. $x_\text{baseline}$ is optimized such that the lateral distance of the center of rotation to the machine $y_\text{min}$ is minimal, i.e. the tightest turn is achieved.}
    \label{fig:steering}
\end{figure}

The four steering joints of HEAP can be individually controlled, which allows for four different steering modes, i.e., crab, front-wheel, rear-wheel and four-wheel steering. The crab steering mode is rather simple. All steering angles are identical, which will allow a diagonal motion without turning. Front, rear and four-wheel steering are implemented using the Ackermann steering principle, where all wheel axes point to a common center of rotation. The baselines for front and rear-wheel steering go through the non-steerable rear and front wheels, respectively. In the case of four-wheel steering, the baseline is not strictly defined. Furthermore, hip ad-/abduction joints also influence the steering angle of the wheels. Figure \ref{fig:steering} shows a configuration with zero ad-/abduction angles on the left and front ad-/abduction joints moved to the right in the right part. Note that both configurations are maximum steering angles as the left wheel on the left and the right wheel on the right are at their steering limits. The distance of the baseline to the robot center $x_\text{baseline}$ is optimized such that the lateral distance of the center of rotation to the machine $y_\text{min}$ is minimized and thus allows for a minimal turning radius.

\subsection{Chassis Controller}
Besides steering, the leg's kinematic arrangement of a walking excavator allows the adjustment of two DOF's at the hip in the ab-/adduction and flexion/extension direction, see Figure \ref{fig:joints}. In order to optimally balance the chassis while moving over uneven terrain, we use its mobility to adapt the legs to the ground topology and simultaneously adjust the height and roll/pitch angle of the cabin. While considering measurements from all leg joints, the control algorithm only regulates the forces in the flexion/extension actuators. We are formulating an optimization problem \cite{Hutter2017Balance} whereby the machine is modeled as a quasi-static system. This is a valid assumption, since the motion is generally slow and gravitational forces are clearly dominant.

The force optimization is paired with a virtual model controller to control height, pitch and roll with virtual forces and torques. The four legs of HEAP allow for changing the contact force distribution without any motion of the chassis. Thus the contact force distribution is optimized to yield minimal contact forces/as equally distributed contact forces as possible while considering force saturation and a minimal contact force to guarantee that all wheels are always in contact. The resulting contact forces are subsequently transformed into cylinder forces.

The chassis controller has been successfully tested on slopes with gradients up to 100\% without any failure or undesired behavior. Thus the introduced simplifications are tolerable. Refer to \cite{Hutter2017Balance} for more details on the implementation.\footnote{\url{https://www.youtube.com/watch?v=5_Eq8CxKkvM&t}}

\subsection{Arm Controller}
The hierarchical optimization by Bellicoso et al. \cite{Bellicoso2016}, which solves the control problem as a constrained quadratic optimization, is used to implement an inverse dynamic controller as well as an inverse kinematic controller. Various hydraulic-specific limits are included as inequality constraints in this framework.  

In the following, the generalized joint positions for the arm (a subset of the generalized coordinates of the excavator) are denoted as $\boldsymbol{q}$ and the generalized joint velocities as $\boldsymbol{u} = \dot{\boldsymbol{q}} \in \mathbb{R}^n$. The generalized forces $\boldsymbol{\tau}\in\mathbb{R}^n$ are related to the actuator forces $\boldsymbol{\tau}_p$ through the diagonal force mapping matrix $E(\boldsymbol{q})\in\mathbb{R}^{n\times n}$ as

\begin{equation} 
  \boldsymbol{\tau} = E(\boldsymbol{q}) \boldsymbol{\tau}_p
\end{equation}

\subsubsection{Inverse Dynamic Hierarchical Optimization} \label{sec:invdyn}
Table \ref{tab:joint_actuation} shows that only 5 joints of the arm are force controllable, because the Rototilt R4 only allows velocity inputs. Thus, for inverse dynamic control it is $n=5$. Those 5 degree's of freedom have one redundant degree which results in 4 task space coordinates ($m=4$) that are controllable. The task space coordinates are chosen as 

\begin{equation}
  \boldsymbol{r} = \begin{bmatrix}
  x_t &
  z_t &
  \theta_t &
  \psi_t
  \end{bmatrix}^T \in \mathbb{R}^m,
  \label{eq:x}
\end{equation}

where the position $x_t$ and $z_t$ and orientation $\theta_t$ of the shovel is represented in the moving cabin frame and the rotation $\psi_t$ of the cabin is w.r.t. the world frame.

The set of task-space forces is analogous to the task-space coordinates defined as
\begin{equation}
  \boldsymbol{f}_t = \begin{bmatrix}
  f_{x_t} &
  f_{z_t} &
  f_{\theta_t} &
  f_{\psi_t}
\end{bmatrix}^T \in \mathbb{R}^m.
\end{equation}
Other task-space forces and moments can not be actuated and are dropped for the sake of simplicity.

The solution vector $\boldsymbol{x}_d$ of the optimization consists of a desired motion $\dot{\boldsymbol{u}}_d$ and desired forces $\boldsymbol{\tau}_d$ and is
\begin{equation}
  \boldsymbol{x}_d = \begin{bmatrix} \dot{\boldsymbol{u}}^T_d & \boldsymbol{\tau}^T_d\end{bmatrix}^T.
\end{equation}
The set of prioritized tasks used in this work are in decreasing priority:\\

\noindent
\begin{minipage}[c]{0.5\linewidth}
    \begin{enumerate}
      \item Equations of motion
      \item Pump flow limit
      \item Cylinder force limits
      \item Cylinder velocity limits
      \item Cylinder position limits
    \end{enumerate}
\end{minipage} 
\begin{minipage}[c]{0.49\linewidth}
\begin{enumerate}
  \setcounter{enumi}{5}
      \item Self-collision avoidance
      \item End-effector orientation
      \item End-effector position
      \item Joint motion damping
\end{enumerate}
\end{minipage}\\
\\
Note that the end-effector motion task is split into two tasks, i.e., orientation and position task of the end-effector. This allows for the orientation task to be of higher priority than the position task. This setup guarantees that if collision avoidance already restricts the end-effector motion, the bucket orientation is treated with higher priority than the translation. Thus, avoiding a collision cannot change the orientation but only the translation of the shovel and there will be no soil spillage.

\newcounter{counterDyn}
\newcommand\showcounterDyn{\stepcounter{counterDyn}\thecounterDyn) }
\textit{\showcounterDyn Equations of Motion Task:} The relation between generalized accelerations $\dot{\boldsymbol u}$ and generalized forces $\boldsymbol\tau$ is defined by the equations of motion (EoM) as
\begin{equation}
\begin{bmatrix} -M & \mathbb{I}_{n\times n} \end{bmatrix} \boldsymbol{x}_d = \boldsymbol{b} + \boldsymbol{g} + J^T \boldsymbol{f}_t^\text{des} + {F}_v \boldsymbol{u} + {F}_s \text{sign}(\boldsymbol{u}),
\end{equation}
where $M(\boldsymbol{q})\in \mathbb{R}^{n\times n}$ is the mass matrix, $\boldsymbol{b}(\boldsymbol{q},\boldsymbol{u}) \in \mathbb{R}^n$ is the vector of centrifugal and Coriolis terms, $\boldsymbol{g}(\boldsymbol{q}) \in \mathbb{R}^n$ is the vector of gravity terms, ${F}_v \in \mathbb{R}^{n\times n}$ is the diagonal viscous friction matrix and ${F}_s \in \mathbb{R}^{n\times n}$ is the diagonal static friction matrix. $\boldsymbol{f}_t^\text{des}$ is the desired end-effector force in task space and is mapped to generalized forces in joint space by the spatial shovel origin Jacobian $J \in \mathbb{R}^{m\times n}$.

\textit{\showcounterDyn Pump Flow Limit:} The hydraulic pump of the excavator can deliver a maximum oil flow of $Q_{\text{max}}$ . 
The flow per cylinder is calculated from the cylinder areas $A_A$ and $A_B$ with $A_A > A_B$. The vector of relevant cylinder areas $A \in \mathbb{R}^{1\times n}$ produce the summed up flow of all cylinders when multiplied with piston velocities and is derived per joint $i$ as

\begin{equation}
    A(i)= 
\begin{cases}
    A_A,& \text{if } \dot{\boldsymbol{q}}(i)>0\\
    -A_B,              & \text{otherwise.}
\end{cases}
\end{equation}

The inequality constraint to avoid exceeding the maximum flow is
\begin{equation}
\begin{bmatrix} \delta t A E(\boldsymbol{q}) & 0
\end{bmatrix}
\boldsymbol{x}_d \leq Q_{\text{max}} - A E(\boldsymbol{q}) \dot{\boldsymbol{u}}
\end{equation}

with $E(\boldsymbol{q})$ being the joint position-dependent transformation from joint velocities to piston velocities.

\textit{\showcounterDyn Cylinder Force/Torque Limits Task:} The hydraulic actuators of the excavator have limited forces/torques $[\boldsymbol{\tau}_{p}^\text{min}, \boldsymbol{\tau}_{p}^\text{max}]$ that they can apply to the system. These limits are configuration dependent and incorporated in an inequality as
\begin{equation}
\begin{bmatrix} 0 & \mathbb{I}_{n\times n} \\ 0 & -\mathbb{I}_{n\times n} \end{bmatrix} \boldsymbol{x_d} \leq \begin{bmatrix} E(\boldsymbol{q})\boldsymbol{\tau}_{p}^\text{max} \\ -E(\boldsymbol{q})\boldsymbol{\tau}_{p}^\text{min} \end{bmatrix}.
\end{equation}

\textit{\showcounterDyn Cylinder Velocity Limits Task:} The maximum flow of oil through a valve limits the maximum velocity a hydraulic cylinder can achieve. The corresponding piston velocity limits $[\boldsymbol{v}_\text{min}, \boldsymbol{v}_\text{max}]$ are transformed to joint velocity limits using the inverse of $E(\boldsymbol{q})$ and recomputed every time step. The inequality constraint is
\begin{equation}
\begin{bmatrix} \delta t \mathbb{I}_{n\times n} & 0 \\ -\delta t \mathbb{I}_{n\times n} & 0 \end{bmatrix} \boldsymbol{x_d} \leq \begin{bmatrix} E(\boldsymbol{q})^{-1}\boldsymbol{v}_\text{max}-\boldsymbol{u} \\ -(E(\boldsymbol{q})^{-1}\boldsymbol{v}_\text{min}-\boldsymbol{u}) \end{bmatrix}.
\end{equation}

\textit{\showcounterDyn Cylinder Position Limits Task:} The hydraulic cylinders also have a limited stroke and therefore the joint positions are limited by incorporating
\begin{equation}
\begin{bmatrix} \frac{\delta t^2}{2} \mathbb{I}_{n\times n} & 0 \\ -\frac{\delta t^2}{2}\mathbb{I}_{n\times n} & 0 \end{bmatrix} \boldsymbol{x_d} \leq \begin{bmatrix} \boldsymbol{q}_\text{max}-\boldsymbol{q}-\boldsymbol{u}\delta t \\ -(\boldsymbol{q}_\text{min}-\boldsymbol{q}-\boldsymbol{u}\delta t) \end{bmatrix},
\end{equation}
where $\boldsymbol{q}_\text{min}$ and $\boldsymbol{q}_\text{max}$ are the joint position limits computed from the actuator position limits.

\textit{\showcounterDyn Self Collision Avoidance Task:} Local self collision avoidance for convex objects is implemented as proposed by Faverjon and Tournassoud \cite{Faverjon_ICRA1987_collision}. The approach limits the velocity of the bodies in direction of the collision. The links are represented with collision primitives, i.e. boxes, and checked for collision using the Bullet physics engine by Coumans et al. \cite{coumans2013bullet}. Bullet will provide the two closest points $\boldsymbol{p}_1$ and $\boldsymbol{p}_2$ between two bodies. The normalized direction of collision is
\begin{align}
    \boldsymbol{n} = (\boldsymbol{p}_2-\boldsymbol{p}_1)/d &&
    \text{with} &&
    d=||\boldsymbol{p}_2-\boldsymbol{p}_1||.
\end{align}
The distance where the constraint will have an influence is $d_i$ and the minimum safety distance is $d_s$. The constraint, that limits the end effector velocity in direction of the collision, is then  
\begin{equation}
\begin{bmatrix} 
\delta t \boldsymbol{n}^T J_{\boldsymbol{p}_1} & 0
\end{bmatrix}
\boldsymbol{x}_d \leq \xi \frac{d-d_s}{d_i-d_s} - \boldsymbol{n}^T J_{\boldsymbol{p}_1}\boldsymbol{u}.
\end{equation}

\textit{\showcounterDyn End-Effector Orientation Task:} A desired end-effector rotational motion $\ddot{\boldsymbol{r}}_{R,\text{des}}$ can be incorporated with the equality
\begin{equation}
\begin{bmatrix} J_R & 0 \end{bmatrix} \boldsymbol{x}_d = \ddot{\boldsymbol{r}}_{R,\text{des}} -{\dot{J}_R}^T \boldsymbol{u}
\end{equation}
using the rotational Jacobian $J_R$ where $J = \begin{bmatrix} J_T & J_R \end{bmatrix}^T$ and the end-effecotr motion $\ddot{\boldsymbol{r}}_{\text{des}} = \begin{bmatrix} \ddot{\boldsymbol{r}}_{T,\text{des}} & \ddot{\boldsymbol{r}}_{R,\text{des}} \end{bmatrix}^T$.

\textit{\showcounterDyn End-Effector Position Task:} A desired end-effector translational motion $\ddot{\boldsymbol{r}}_{T,\text{des}}$ can be incorporated with the equality
\begin{equation}
\begin{bmatrix} J_T & 0 \end{bmatrix} \boldsymbol{x}_d = \ddot{\boldsymbol{r}}_{T,\text{des}} -{\dot{J}_T}^T \boldsymbol{u}
\end{equation}
using the translational Jacobian $J_T$.

\textit{\showcounterDyn Joint Motion Damping Task:} If there is still a nullspace left, the controller dampens the joint motion through the equality task
\begin{equation}
    \begin{bmatrix} \mathbb{I}_{n\times n} & 0 \end{bmatrix} \boldsymbol{x}_d = - k_p \boldsymbol{u}.
\end{equation}

\subsubsection{Inverse Kinematic Hierarchical Optimization}\label{sec:invkin} 

The same hierarchical optimization framework can be used to implement an inverse kinematic controller for the arm which does not have any notion of force but also does not need an accurate dynamic model. Inverse kinematics for manipulators exploiting the redundancy was first shown by Hanafusa et al. \cite{Hanafusa1981}. As shown in table \ref{tab:joint_actuation}, all the arm joints can be velocity controlled. Thus, $n=7$ and $m=6$ for the inverse kinematic controller means that a full 6 DOF task space pose of the end-effector can be controlled. The solution of the optimization is the desired joint motion $\boldsymbol{u_d}$. The set of prioritized tasks are in decreasing priority:

\newcounter{counterKin}
\newcommand\showcounterKin{\stepcounter{counterKin}\thecounterKin) }
\textit{\showcounterKin Pump Flow Limit Task:}
\begin{equation}
A E(\boldsymbol{q}) \boldsymbol{u_d} \leq Q_{\text{max}}
\end{equation}

\textit{\showcounterKin Cylinder Velocity Limits Task:}
\begin{equation}
\begin{bmatrix} \mathbb{I}_{n\times n}\\ -\mathbb{I}_{n\times n} \end{bmatrix} \boldsymbol{u_d} \leq \begin{bmatrix} E(\boldsymbol{q})^{-1}\boldsymbol{v}_\text{max} \\ -(E(\boldsymbol{q})^{-1}\boldsymbol{v}_\text{min}) \end{bmatrix}
\end{equation}

\textit{\showcounterKin Cylinder Position Limits Task:}
\begin{equation}
\begin{bmatrix} \delta t \mathbb{I}_{n\times n}\\ -\delta t \mathbb{I}_{n\times n}\end{bmatrix} \boldsymbol{u_d} \leq \begin{bmatrix} \boldsymbol{q}_\text{max}-\boldsymbol{q}-\boldsymbol{u}\delta t \\ -(\boldsymbol{q}_\text{min}-\boldsymbol{q}-\boldsymbol{u}\delta t) \end{bmatrix}
\end{equation}

\textit{\showcounterKin Self Collision Avoidance Task:}
\begin{equation}
\boldsymbol{n}^T J_{\boldsymbol{p}_1} \boldsymbol{u}_d \leq \xi \frac{d-d_s}{d_i-d_s}
\end{equation}

\textit{\showcounterKin End-Effector Orientation Task:} A desired end-effector rotational motion $\dot{\boldsymbol{r}}_{R,\text{des}}$ can be incorporated with the equality
\begin{equation}
J_R \boldsymbol{u}_d = \dot{\boldsymbol{r}}_{R,\text{des}}.
\end{equation}

\textit{\showcounterKin End-Effector Position Task:} A desired end-effector translational motion $\dot{\boldsymbol{r}}_{T,\text{des}}$ can be incorporated with the equality
\begin{equation}
J_T \boldsymbol{u}_d = \dot{\boldsymbol{r}}_{T,\text{des}}.
\end{equation}

\textit{\showcounterKin Minimize Cylinder Velocities Task:} If there is still a nullspace left, the controller minimizes the joint velocities
\begin{equation}
\min\|\boldsymbol{u_d}\|_2.
\end{equation}

The desired end-effector velocity $\dot{r}_{\text{des}}$ is computed from the desired end effector trajectory defined by a velocity $\dot{r}_{\text{traj}}$ and a position $\boldsymbol{r}_{\text{traj}}$, as
\begin{equation}
    \dot{r}_{\text{des}} = \dot{r}_{\text{traj}} + \text{PID}(\boldsymbol{r},\boldsymbol{r}_{\text{traj}})
\end{equation}
with position feedback through a PID controller. Both $\dot{r}_{\text{traj}}$ and $\boldsymbol{r}_{\text{traj}}$ are extracted from cubic Hermite splines.

\section{Applications}\label{sec:applications}
\begin{figure*}[bt]
    \captionsetup[subfigure]{justification=centering}
    \centering
    \subcaptionbox{Autonomous excavation of free-form embankments (June 2020).\\ Video: \url{https://www.youtube.com/watch?v=Wjq3Nf9rWrM}
    \label{fig:landscaping}}[.49\linewidth]{%
        \includegraphics[width=0.99\linewidth]{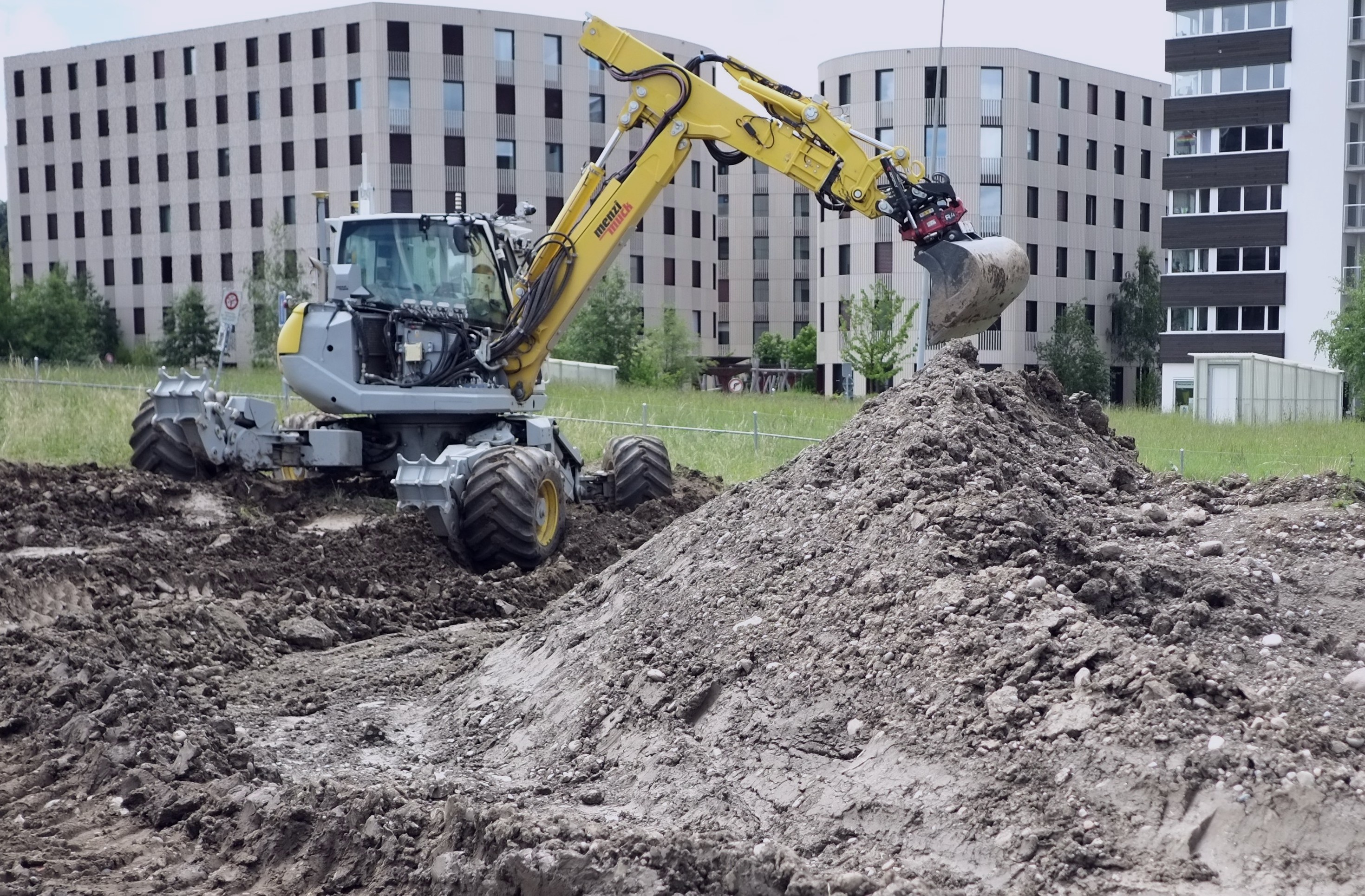}%
    }
        \hfil
    \subcaptionbox{Autonomous assembly of dry stone walls (August 2020).\\ Video: \url{https://youtu.be/1R1jaulXitg}
    \label{fig:stone_aggregation}}[.49\linewidth]{%
        \includegraphics[width=0.99\linewidth]{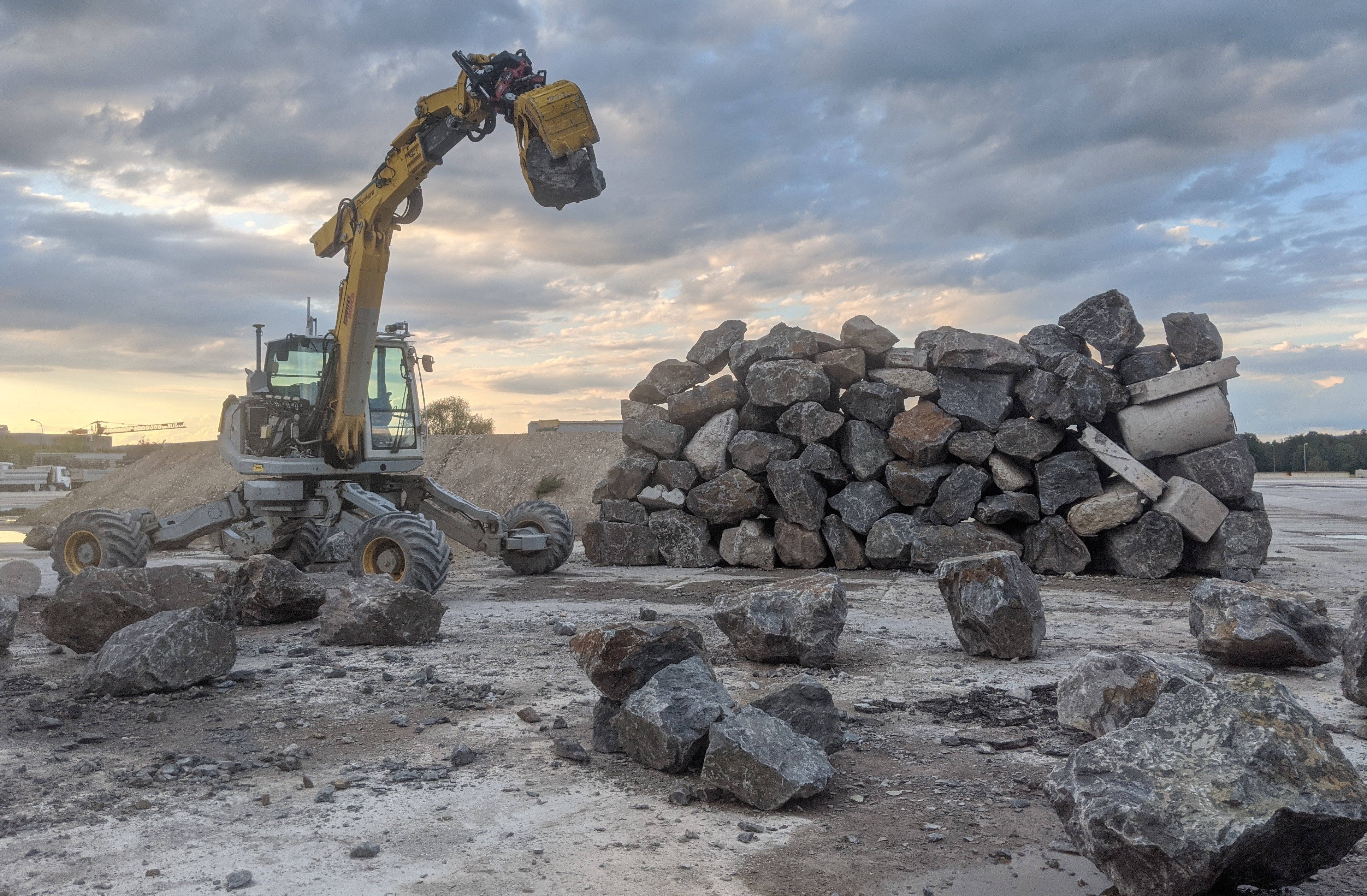}%
    }
        \hfil
    \subcaptionbox{Autonomous forestry work (October 2019).\\ Video: \url{https://youtu.be/NZAXnmpPR7c} 
    \label{fig:forestry}}[.49\linewidth]{%
        \includegraphics[width=0.99\linewidth]{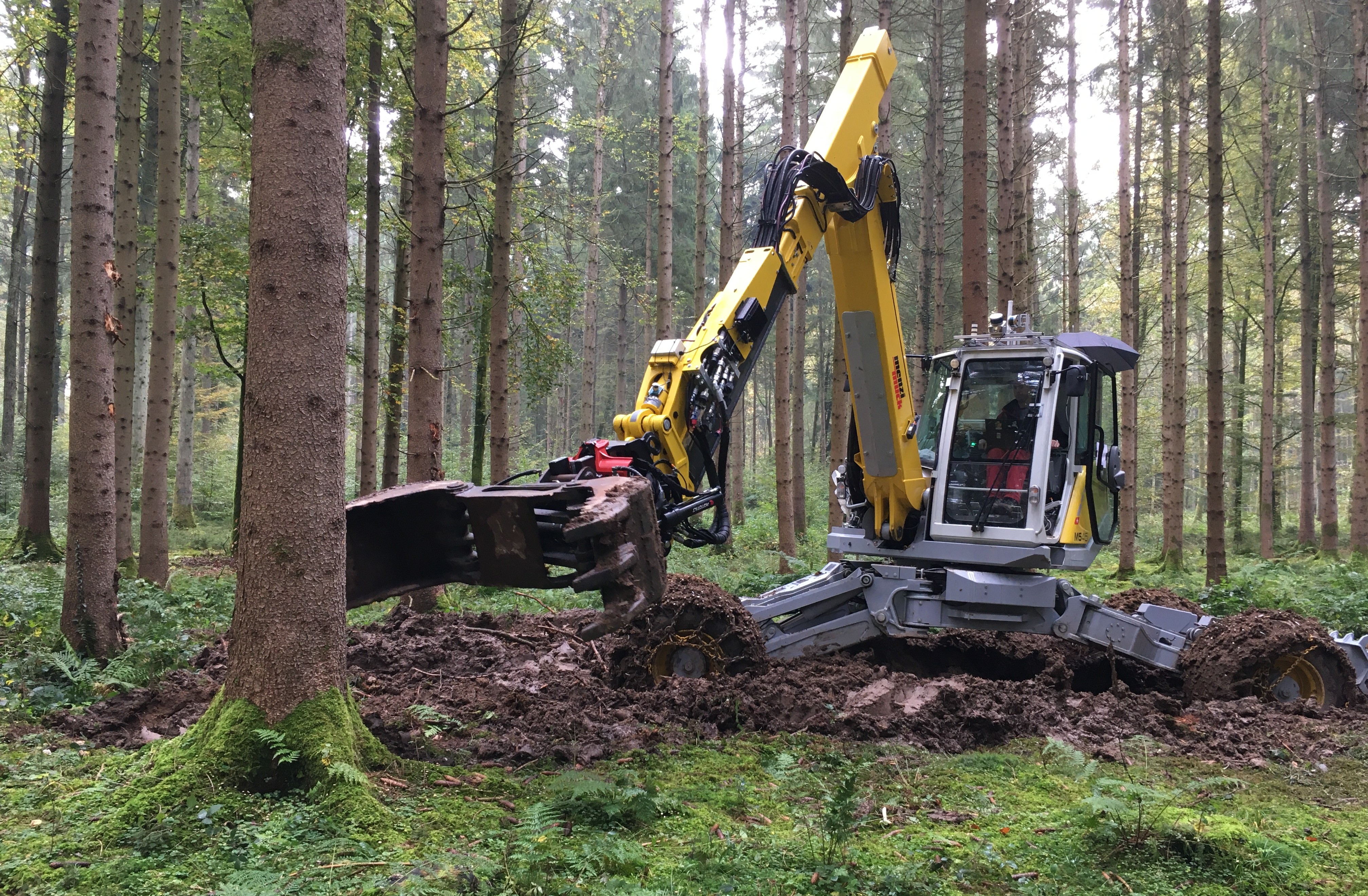}%
    }
        \hfil
    \subcaptionbox{Semi-autonomous teleoperation for cleaning up rockslides (August 2020) and excavating buried ammunition (January 2020).\\ Video: \url{https://www.youtube.com/watch?v=IbMZTErlQNU}
    \label{fig:teleoperation}}[.49\linewidth]{%
        \includegraphics[width=0.99\linewidth]{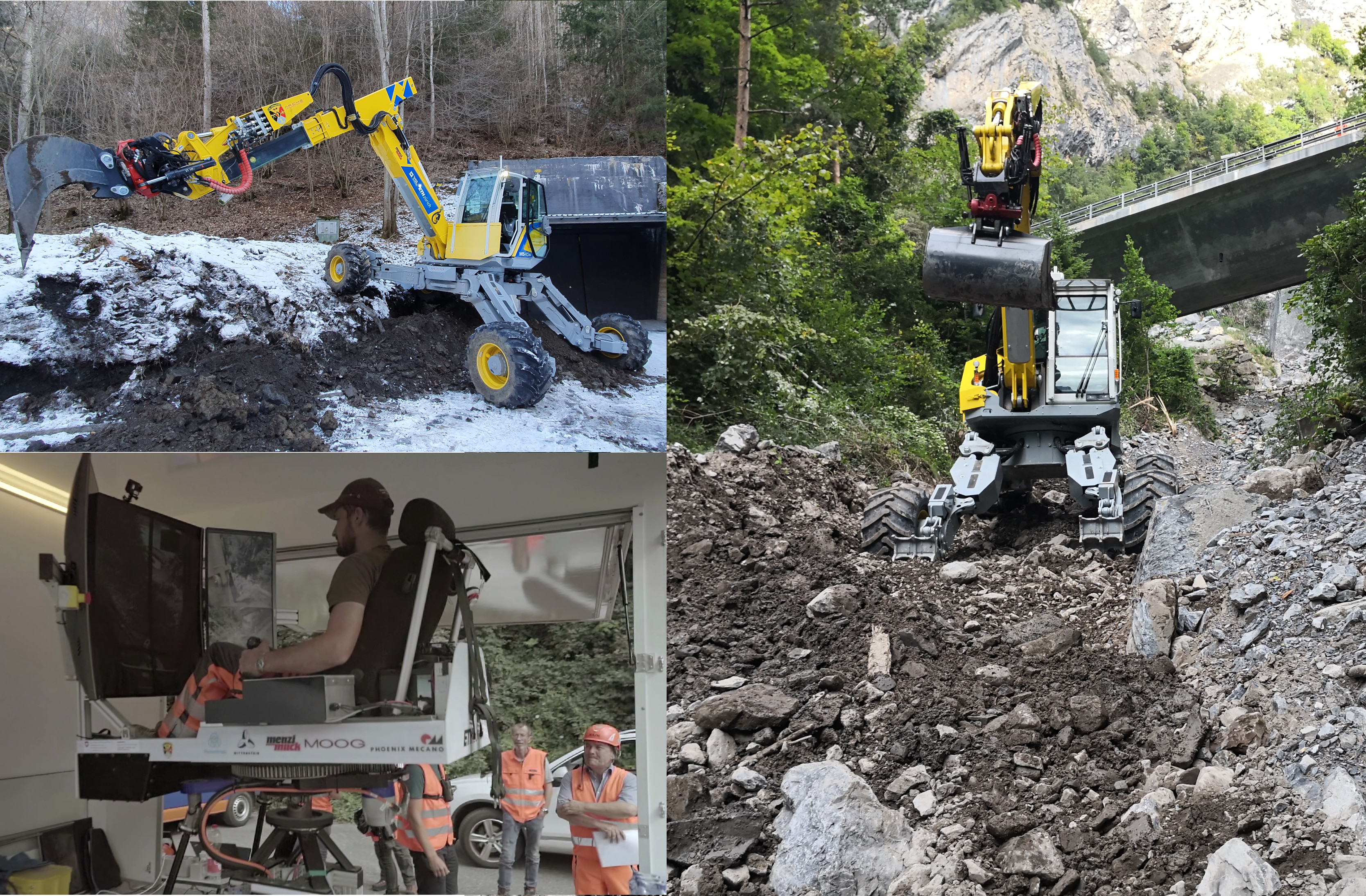}%
    }
    \caption{HEAP has been deployed in many different autonomous scenarios, e.g. (\protect\subref{fig:landscaping}) excavation, (\protect\subref{fig:stone_aggregation}) dry stone wall assembly and (\protect\subref{fig:forestry}) forestry work, as well as (\protect\subref{fig:teleoperation}) semi-autonomous teleoperation. For all applications, operators are only present in the cabin for safety and do not interact with the machine.}
    \label{fig:applications}
\end{figure*}

The work shown in this article lays the foundation for HEAP's use in various missions. Figure \ref{fig:applications} provides an overview of four example applications: autonomous excavation of free-form embankments, autonomous assembly of dry stone walls, autonomous forestry work and semi-autonomous teleoperation in dangerous situations. Note that there are videos attached to each application in Figure \ref{fig:applications} for better illustration.

\subsection{Autonomous Excavation}
HEAP was first used to autonomously and precisely excavate embankments with free-form shapes in June 2020. The work built upon our previous work on free-form trenches \cite{Jud2017} \cite{Jud2019}. The chassis controller guarantees a stable stance of the machine for high precision excavation. The inverse dynamic hierarchical optimization is used for force-controlled digging and the inverse kinematic counterpart for motions in the air, e.g., dumping soil. The proposed state estimator is used to position the bucket accurately. The excavation-specific planner provides task space digging forces and dump locations for the ID and IK arm controllers respectively, as well as the desired base orientation for the chassis controller and a desired base location for the driving controller. With this setup, a two-faced embankment with $0.03\unit{m}$ average error and a more complex s-curved embankment (\SI{30}{\cubic\meter} of soil displaced) with $0.05\unit{m}$ average error can be created autonomously.

\subsection{Autonomous Assembly of Dry Stone Walls}
HEAP can also be operated with a two-finger gripper, instead of a traditional bucket. The gripper is instrumented with an encoder to measure the opening angle. In combination with the inverse kinematic arm controller, which uses all 7 DOF's of the arm, complex objects can be manipulated. This is used to build dry stone walls from stones with irregular shapes. The stones are mapped by the machine itself, which picks them up and moves them in front of the roof-mounted \ac{LiDAR}'s. The accuracy of the generated meshes of stones benefits from the high end-effector accuracy when using the \ac{IMU}'s installed on the arm for joint sensing. The task-specific planner provides grasping poses for the stones and a final pose in the wall with a feasible approach trajectory. A $4\unit{m}$ tall, $10\unit{m}$ long and $2\unit{m}$ wide wall was successfully built with 109 stones from a collection of 131 scanned stones. The stones were a mixture of natural stones and concrete blocks. They had an average weight of $1030\unit{kg}$ (min. $445\unit{kg}$, max. $2425\unit{kg}$) and were placed with a mean error of only $0.128\unit{m}$. Autonomously placing a single stone took approximately $700\unit{s}$ or $1300\unit{s}$, if the stone had to be flipped. \cite{Johns2020}

\subsection{Autonomous Forestry Work}
In a case study on autonomous forestry, HEAP was deployed to harvest trees. Due to an unreliable \ac{GNSS} connection in dense forests, an additional \ac{LiDAR} was used for localization. Additionally, the exposed draw-wire encoders on the arm could have easily been harmed by branches of nearby trees and were removed. Thus only \ac{IMU}'s were used for joint state estimation. The chassis controller allowed the machine to drive over large tree stumps and in deep muddy terrain while keeping all four wheels in contact for stability and reducing ground compaction. A task-specific planner provided grasp poses and feasible approach trajectories to the trees for the IK arm controller and steering angles and driving speed for the driving controller to autonomously move through the forest. The chassis was commanded to stay level over the entire mission. Multiple trees were successfully grabbed in a \SI{100}{\meter} by \SI{30}{\meter} forest. It took approximately \SI{3}{\minute} to navigate to a tree and grab it, where the gripper could be placed with less than \SI{30}{\centi\meter} error and the chassis with less than \SI{40}{\centi\meter} error. 

\subsection{Semi-autonomous Teleoperation}
Because HEAP was successful in these autonomous missions, we built a second, identical machine for semi-autonomous teleoperation. Equipped with multiple cameras, it was used to dig a trench in soil containing live World War II ammunition, where a standard approach with an operator in the cabin would be too dangerous. The front of the machine was nonetheless armour-plated with steel and explosion-proof glass to prevent a total loss of the machine in a worst-case scenario. Low-latency video streaming of one 4K and four 2.5K cameras simultaneously was achieved with a latency of only $100\unit{ms}$. A $200\unit{m}$ trench to lay a cable was successfully dug while bringing up a large amount of ammunition without jeopardizing any workers. The semi-autonomous teleoperated excavator was also used to clean rocks out of a steep creek in a dangerous rock fall area. Without this periodic maintenance, a road and train bridge would be at risk from the rocks. The balancing chassis is vital during teleoperation in rough terrain. The remote operator lacks the sense to keep all four legs in contact, which might result in the machine falling over. With the actively controlled chassis, the remote operator can focus on the work with the arm and the machine handles the legs. The automatic chassis was well received by experienced operators, and considered irreplaceable for teleoperation.

\section{Development Insights} \label{sec:lessons_learned}

A few insights gathered during HEAP's development are summarized below. We also offer practical advice on how to address these problems:

\textit{A regular excavator is not a high precision robot.} Instead, it is built to last for decades in the roughest environments. Thus, joints have play, which gets worse the older the machine is. The oil quality of these machines can be challenging for high-performance hydraulic parts, such as the nozzle-flapper pilot stage clogging up in the first generation servo valves. Furthermore, excavators are built from steel and are exposed to high temperature ranges, which has a significant effect on link lengths. Thus, it is crucial to select hydraulic parts and sensors with these circumstances in mind. For example, joint angle sensors can provide better end-effector accuracy than cylinder length sensors for any manipulation task, i.e. excavation, tree harvesting, stone stacking. If an even higher accuracy is needed, a localization module, e.g., as shown by Gawel et al. \cite{Gawel2019}, can be placed on the end-effector itself, compensating for any inaccuracies of the arm. 

\textit{The installed hydraulic pump and the corresponding controller are not meant for servo valves.} Thus installing servo valves in an excavator is not as straightforward as simply replacing the regular main valves. Instead, the entire hydraulic system has to be considered. The faster servo valves can cause pressure drops and peaks that the pump and its controller are too slow to react to. This cannot be solved by retuning the pump controller for this specific use case, because if the pump is too aggressive, the engine will be the slowest element in the chain and it will stall. Instead, an accumulator on the supply line can decouple the valves from the drive train to a certain degree, as used in this work. An ideal solution would be an arm controller that incorporates a model of the hydraulic power unit. However, as precisely modelling these parts is difficult, we need to think about novel approaches to improve control performance with standard components, such as our reinforcement learning approach \cite{Egli2020}.

\textit{Hydraulics does not result in a rigid and "stiff" system.} Often, hydraulics is seen as being rigid and fairly stiff compared to other actuation principles. This is not the case. In the setup introduced in this article, where the servo valves for the arm are located next to the cabin and connected to the respective cylinders with long hoses, pressure oscillations between valve and cylinder can be easily provoked. Fluid compressibility and elastic expansion of hoses, especially during high ambient temperatures, cause these issues. This was especially apparent during velocity-controlled motions in the air using servo valves, as there would be no measurable/visible motion of the cylinders but high oscillations in the pressure measurements. Regular excavator hydraulics do not suffer as intensively, because the pressure dynamics is an order of magnitude faster than control bandwidths in conventional machines. In this work, the frequencies causing these issues are removed from the feedback signal with a notch filter. Another possible solution would be to rigorously model all the components. However, this quickly becomes unfeasible for highly complex machines.

\textit{The weight of the machine does not guarantee a stable stance.} Excavators are heavy and slow. However, this does not imply that they have a stable stance. Specifically, on a walking excavator with air filled tires, the tires act like springs and unfortunately, the control bandwidth of the arm, when being moved in the air, is close to the natural frequency of this rocking motion. Similar effects can also be seen on tracked excavators on soft ground. This is not an issue if the arm is in ground contact, as the machine's support area is increased by the additional contact. A benefit of the active chassis is that the virtual model controller can be tuned to dampen these motions actively. For other machines without the benefit of an active chassis, a control approach could be used, e.g., a frequency-aware model predictive control \cite{Grandia2019} that avoids exciting these frequencies  with the arm. Overall, the active chassis has shown itself to be absolutely critical for any kind of unmanned operation. It guarantees that all four wheels are always in ground contact. This is not a given for other machines without an active chassis, which could be prone to tipping over when moving the excavator's arm.

\section{Conclusion and Outlook} \label{sec:conclusion}

This article introduced HEAP, the autonomous walking excavator. Its main contributions are: A detailed system description of sensors and actuators that allows a replication of the platform. Servo valves, electric pilot stage driven main valves and a novel actuated joystick are all implemented together on one machine allowing for a meaningful comparison. Furthermore, the upgrade to revision 2 actuators in the chassis is described. A state estimation approach for wheeled legged robots is proposed and evaluated for the specific use case with excavators where the wheels tend to slip often laterally and longitudinally. Controllers for driving the chassis and the arm  are also described. These contributions are the foundation to execute advanced autonomous missions for excavation, building dry stone walls or forestry work. In these unique autonomous missions, HEAP has shown unseen end-effector accuracy and unprecedented manipulation skills, e.g., in excavating free-form 3D landscapes or accurately placing large rocks to build a wall.

So far, the joint level hydraulic control uses traditional methods and a hand-tuned controller that neglects complex nonlinear hydraulic effects and coupling. This could be vastly improved with a more advanced control method that does not neglect the highly nonlinear dynamic behaviour of hydraulic systems. To overcome this limitation, we have started to work on learning algorithms for control. These learning based approaches are expected to outperform traditional control methods in the future, possibly improving the pilot stage driven main valves enough to make the costly servo valves obsolete. A second point of improvement is the state estimation. We have shown that legged odometry and rolling prediction might not be beneficial to use with this particular machine and scenarios. However, using \ac{GNSS} \ac{RTK} and \ac{IMU} only, heavily relies on the connection reliability of the \ac{GNSS} \ac{RTK}. The state estimator can be made more robust by adding visual odometry or localization. The residual based filter would allow for straightforward integration.

Future development will improve the machine in various areas. A custom built force torque sensor at the end-effector will allow more precise manipulation of objects and better control of interaction forces to facilitate more challenging manipulation tasks. Furthermore, a whole body controller handling the chassis and arm in one instance instead of the individual controllers shown in this work would allow advanced manoeuvres reserved for walking excavators, e.g. stepping over obstacles using the arm \cite{Jelavic2020}. However, these motions have to be planned first. Discovering motion plans with complex contact schedules alternating between wheel and arm contacts to overcome gaps, climb up steps or reposition the machine will be a topic of future research. Further development is especially important for instances when these plans have to be discovered quickly (in just a few seconds) to not negatively affect the machine productivity. 

The autonomous excavator presented in this article has shown high manipulation accuracy in various tasks. The absolute accuracy and measuring capabilities of such a robot are also where future applications' benefits will lie. In autonomous excavation, the machine knows exactly how much soil it is moving, and also when and where. This allows material-neutral designs to be created on construction sites, removing the necessity to bring or remove excess soil and drastically improving sustainability. The potential for absolute accuracy will also lead to new design possibilities and construction methods, e.g., a vault supported during construction by an accurately shaped earth pile. 

Although autonomous excavators already show higher absolute accuracy in many tasks compared to human operators, there are still things it cannot do as well. When a task needs constant adaptation and shows many unexpected events, a human operator can still reason and leverage his experience to solve the problem where a robot would simply be stuck. Repetitive and well-controlled tasks, e.g., digging a many kilometer-long pipeline trench through a desert, will be autonomously executed by robots soon, but as the research in this article is developed further, autonomous walking robots will begin to be able to tackle more and more sophisticated excavation and construction projects.

\bibliography{bib/application,bib/control,bib/introduction,bib/state_estimation,bib/system_description}

\end{document}